\documentclass[final,1p,times]{elsarticle}

\usepackage{amssymb}

\usepackage{lmodern}        

\usepackage{amsmath}
\usepackage{amsthm}

\usepackage{array}
\usepackage{pifont}
\usepackage{subcaption}
\usepackage{multirow}
\usepackage{etoolbox}
\usepackage{natbib}
\usepackage{color}
\usepackage[dvipsnames]{xcolor}
\usepackage{ulem}
\apptocmd{\sloppy}{\hbadness 10000\relax}{}{}
\usepackage{enumitem}
\usepackage{algorithm}
\usepackage{algpseudocode}
\usepackage{textcomp}
\usepackage{hyperref}
\usepackage{booktabs}
\usepackage{siunitx}

\algrenewcommand\algorithmicrequire{\textbf{Input:}}
\algrenewcommand\algorithmicensure{\textbf{Output:}}

\newcommand{\fetwo}{FE$^\mathrm{2}$}

\begin{document}

\begin{frontmatter}



\title{Microstructure-Conditioned Surrogate Models for Graded Multiscale Optimization of Mycelium Composites}

%

\author[label1]{J. Storm}
\author[label1]{I. B. C. M. Rocha}
\author[label2]{S. Schyck}
\author[label2]{K. Masania}
\author[label1]{F. P. van der Meer}


 \affiliation[label1]{organization={SLIMM Lab, Faculty of Civil Engineering and Geosciences, Delft University of Technology},
             addressline={P.O. Box 5048},
             city={Delft},
             postcode={2600GA},
             state={Zuid-Holland},
             country={The Netherlands}}
 \affiliation[label2]{organization={Shaping Matter Lab, Faculty of Aerospace Engineering, Delft University of Technology},
             city={Delft},
             state={Zuid-Holland},
             country={The Netherlands}}


\begin{abstract}
Emerging sustainable materials increasingly rely on engineered hierarchy and microstructure to achieve control of their properties and mechanical behavior.
Optimizing these materials with controllable microstructures requires efficient multiscale simulations.
Data-driven surrogate models for the microscale can accelerate multiscale simulations, but require large amounts of data even for a fixed microstructure.
When a range of microstructures is considered, as is the case in multiscale optimization, even more data is needed to train a surrogate.
To overcome this challenge, we condition a hybrid physics-data surrogate on microstructural variables using a hypernetwork.
This approach enables accurate predictions of multiscale mechanical behavior for a mycelium-woodchip composite material, even when trained on small datasets.
The conditioned surrogate makes multiscale simulations of functionally graded structures tractable, and we validate it against a full \fetwo{} simulation.
We optimize a graded multiscale disk, and reduce the peak stress by 43\% compared to one with a random microstructure.
Then, we go one step further, conditioning the network directly on manufacturing variables that can have a complex influence on the microstructure.
This is a practical route to engineer the microscale for desired macroscale behavior.
This contribution highlights the benefits of microarchitectured structures and demonstrates how conditioned surrogate models enable their multiscale optimization, which will accelerate the development and design of future sustainable materials and structures.
\end{abstract}




\end{frontmatter}


\section{Introduction}\label{sec1}
Homogeneous structural materials are widespread due to their uniformity and ease of manufacturing.
However, in nature, many heterogeneous materials evolved, such as bones, teeth, wood, and bamboo, as these are more mechanically efficient designs \citep{Miyamoto2013}.
These are examples of functionally graded materials, characterized by having continuous spatial variation of material properties at the macroscale.
Precisely manufacturing a functionally graded material similarly enables the design of more efficient structures.
However, optimizing such a grading requires relating the local microstructure to the macroscopic response.

\fetwo{} is a concurrent approach to simulate both scales, where microscale representative volume elements (RVEs) are embedded in a macroscale domain.
Solving a separate boundary value problem for every quadrature point makes this very computationally demanding.
Instead, data-driven models can act as surrogates for the microscale simulations, estimating the homogenized microscale response at a fraction of the computational costs.
Many surrogates have been developed, both purely data-driven and based on physical principles, capturing the behavior in materials ranging from hyperelasticity and viscoelasticity to plasticity, damage, and fracture \citep{Fuhg2024, Chafia2025, Ammendolea2026}.
For hyperelastic materials in particular, several architectures exist that emphasize satisfying all physical requirements, such as thermodynamic consistency and polyconvexity \citep{Linka2021, Klein2022}.
For linear elastic materials, data-driven approaches have enabled multiscale graded optimization by predicting an effective elasticity tensor from the microstructure \citep{Zheng2021, Kim2021, Chandrasekhar2023}.
For nonlinear materials, however, surrogates are typically trained to map strains to homogenized stresses for a single, fixed microstructure, and therefore cannot account for varying underlying microstructures.
This makes them unsuitable for simulating functionally graded composites, where the microstructure varies across the domain.

One way of overcoming this limitation is by using a surrogate model that takes the complete microstructure as input, for example, using a convolutional neural network (CNN) \citep{Rao2020, Pitz2024} or a graph neural network (GNN) \citep{Storm2024}.
Microstructural geometric variables can also be used to directly predict the effective properties of the material \citep{Black2023}.
Alternatively, they can be passed as additional inputs, for example to the hidden state in a gated recurrent unit \citep{Mozaffar2019}, or encoded into a small vector before concatenating it to the hidden embedding in a network \citep{Jailin2025}.
These approaches have the potential to be used in the optimization of a graded material, but generally still require vast amounts of training data.
Meta-learning or transfer learning can adapt or fine-tune a base model to limited new data from a different underlying distribution \citep{Caruana1994, Finn2017, Heidenreich2024, Ghane2024}, but still require separate data collection for each new setting, making it infeasible for continuous gradings.

An effective approach to reducing the required amount of training data in surrogates is to incorporate a physical bias in the model.
Physically Recurrent Neural Networks (PRNNs) have shown a remarkable ability to learn based on tiny datasets, overcoming the bottleneck of costly data generation \citep{Maia2023}.
A key feature of PRNNs is their modularity, inheriting the properties of the embedded microscopic constitutive models.
For instance, embedding a history-dependent material model yields a history-dependent PRNN.
The material properties of the embedded material model can be modified directly without requiring retraining or additional data collection \citep{Kovcs2025}.
The PRNN is thus effectively conditioned on these material parameters, despite being trained with only a single parameter value.
Recent work has proposed an energy-based PRNN formulation that is thermodynamically consistent \citep{Maia2026}.

In a similar vein, a Deep Material Network (DMN) learns the homogenization behavior of a multi-phase material \citep{Liu2019}, while maintaining the microscale material models.
DMNs have recently been conditioned on the microstructure, such as by interpolating between several DMNs trained on different RVEs, by conditioning on the volume fraction \citep{Li2024}, and by conditioning on a latent space learned from the microscale geometry \citep{Wei2025}.
Since a DMN essentially learns the homogenization scheme, this conditioning can only be done by modifying the weights of the network.
The weights of the DMN are thus predicted by a separate neural network; such a second network is generally called a \textit{hypernetwork}.
HyperCANs follow the same principle, where, based on the geometry of a unit cell truss lattice, a hypernetwork learns the weight parameters of an input convex neural network that predicts the strain energy density given a deformation gradient \citep{Zheng2024}.

\begin{figure}[htbp]
    \centering
    \includegraphics[width=1.\textwidth]{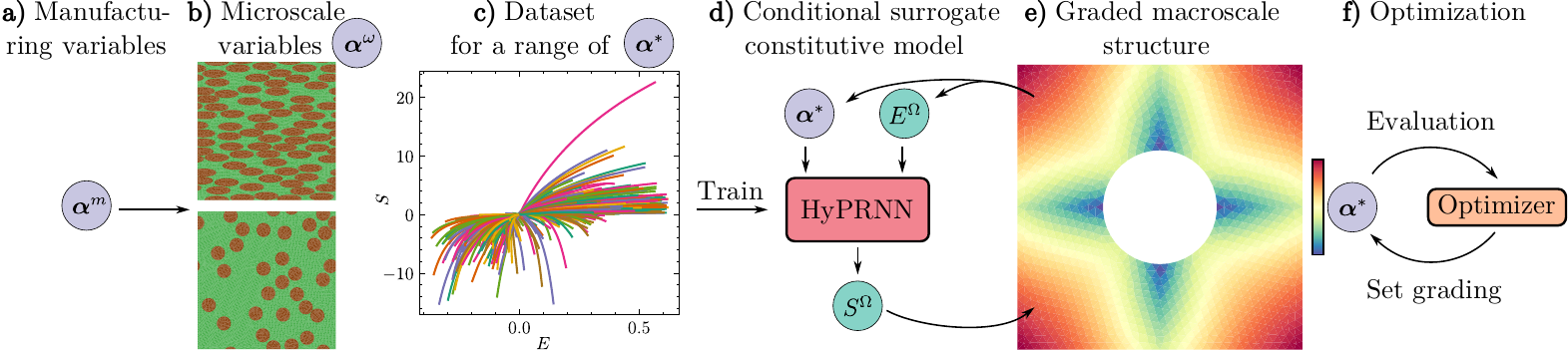}
    \caption{Overview for optimizing a graded multiscale material using a surrogate model conditioned on either microscale variables $\boldsymbol\alpha^{\omega}$ or manufacturing variables $\boldsymbol\alpha^{m}$.}
    \label{fig:overview}
\end{figure}

In this work, we opt for PRNNs because they can already account for variations in the microscale material properties, and extend them to be conditioned on microstructural variables using a hypernetwork, which we refer to as a \textit{HyPRNN}.
A schematic overview of the methods in this work is presented in Figure \ref{fig:overview}.
Our contributions are as follows:

\begin{enumerate}[noitemsep]
    \item \textit{Conditioning a PRNN on microstructural variables.}
    While traditional PRNNs can already account for differences in microstructural material properties, they cannot account for geometric variations in the microstructure.
    We enable this using a hypernetwork that learns the PRNN weights based on the microstructural variables.
    \item \textit{PRNNs for geometric nonlinearity in finite strains.} To capture geometrically nonlinear effects, we introduce a more flexible nonlinear PRNN encoder. In addition, we modify the energy-based PRNN formulation of \citep{Maia2026} to account for large deformations.
    \item \textit{Demonstration of a tractable multiscale optimization.} With the HyPRNN, we iteratively run a graded multiscale simulation inside an optimization algorithm to find a grading that leads to better mechanical performance.  
    \item \textit{Conditioning on manufacturing variables.}
    When the microstructure exhibits a complex geometry that cannot be parametrized explicitly, a common strategy is to condition on a black-box latent space learned by a generative model that captures its properties \citep{Wei2025}.
    We introduce a new approach that instead conditions directly on the manufacturing variables generating the microstructure, eliminating the need for such an additional model.
    This approach is demonstrated by conditioning on the inputs of a discrete-element method simulation that creates the microscopic geometry.
\end{enumerate}

\section{Inspiration: mycelium-woodchip composites}\label{sec:inspiration}
\begin{figure}[htbp]
    \centering
    \includegraphics[width=1.\textwidth]{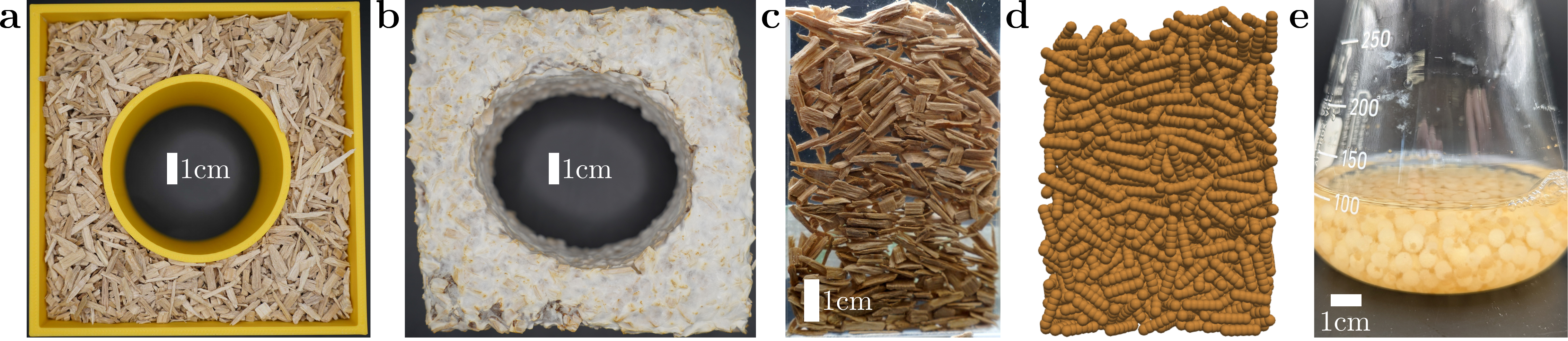}
    \caption{The manufacturing process of a mycelium-woodchip composite. \textbf{a} shows formwork in which woodchips are dropped. After inoculation with mycelium and a few weeks of growth, this becomes a mycelium-woodchip composite sample as shown in \textbf{b}. One method of creating the woodchip distribution is by depositing them in a container. \textbf{c} shows a side-view picture of such a deposition in a thin container, and \textbf{d} shows a discrete-element method based gravity deposition aiming to mimic this process. In \textbf{e} a picture of pure mycelium pellets is shown.}
    \label{fig:fungi_composite}
\end{figure}

While this work focuses on computational methods, we base our requirements on applications such as functionally graded bio-based composites.
In particular, we are interested in mycelium-woodchip composites, which are actively researched for the construction, packaging, and insulation industries \citep{Wang2024}, and benefit from microscale structure.
Mycelium is considered the vegetative root-like structure of fungi and is composed of hyphae, which are tube-like fibers with a diameter of about 1 [$\mu$m].
Growing on a substrate of woodchips or natural fibers, it acts as a binder that forms a lightweight yet stiff biodegradable composite \citep{Gantenbein2022, Schyck2026}.
For instance, Figure \ref{fig:fungi_composite}a-b shows woodchips in a mold, to which inoculated water is added, forming a mycelium-bound composite.
As an engineered living material, its unique properties include the ability to grow and self-repair \citep{McBee2021, Nettersheim2024}, and has been explored for mechanical use \citep{Gantenbein2022, Zhang2023}.
Controlling the substrate properties and mycelium growth can tailor a functionally graded material to specific macroscale behavior.

Modeling this macroscale behavior requires the mechanical properties of the individual constituents.
Mechanical tests on pure mycelium by Islam et al. revealed that its Young's modulus depends on the density, giving an average of $E$ = 1.3 [MPa] for a density of 34 [kg/m$^3$], and Poisson's ratio $\nu$ = 0.275 [-] (Lamé parameters $\lambda$ = 0.62 [MPa], $\mu$ = 0.51 [MPa]) \citep{Islam2017, Islam2018, Islam2018b}.
We adopt these properties and, motivated by their density dependence, also consider a variation in $\mu$.
Mycelium is often characterized similarly to foam, and its constitutive behavior is commonly modeled using a hyperelastic material model \citep{Wang2024, Islam2018, Islam2018b}.
For the woodchips, isotropy and standard beech properties are assumed, with $E$ = 13000 [MPa] and $\nu$ = 0.375 [-].

Various aspects of the manufacturing process of a mycelium-woodchip composite are shown in Figure \ref{fig:fungi_composite}.
Woodchips are inoculated with mycelium spores and placed in formwork, where the mycelium grows over the span of a few weeks to create the composite material.
The growth depends on many factors, such as available nutrients, water content, and air.
A graded material can be formed by, for example, controlling the distribution of both woodchips and pellets. 
Figure \ref{fig:fungi_composite}c,e shows these constituents, and Figure \ref{fig:fungi_composite}d shows the result of numerically simulating the woodchip deposition process.
Experimentally finding the optimal grading of woodchip distribution and angle is not feasible, and even numerically this remains challenging.

\section{Finite strain mechanics}\label{sec:methods}
This section briefly introduces the computational methods and notations associated with finite strain continuum mechanics and hyperelasticity.
For a reference point $\mathbf{X}$ in the initial configuration of a body undergoing deformation, we can define a function $\boldsymbol{\varphi}$ that maps the point to an updated configuration $\mathbf{x}=\boldsymbol{\varphi}(\mathbf{X},t)$ at time $t$.
The deformation gradient of the material follows as
\begin{equation}
  \mathbf{F} = \frac{\partial \boldsymbol{\varphi}}{\partial \mathbf{X}},
  \quad \text{with}
  \quad J=\det \mathbf{F} > 0.
\end{equation}
Commonly used deformation tensors are the right Cauchy-Green tensor ($\mathbf{C}$) and the Green-Lagrange strain ($\mathbf{E}$):
\begin{equation}
  \mathbf{C}=\mathbf{F}^T\mathbf{F}, 
  \qquad \mathbf{E}=\dfrac{1}{2}(\mathbf{C}-\mathbf{I}).
\end{equation}
A hyperelastic solid derives stress from a stored strain energy density $\Psi=\Psi(\mathbf{F})$.
The stress measures are computed as
\begin{equation}
  \mathbf{P}=\frac{\partial \Psi}{\partial \mathbf{F}},
  \qquad
  \mathbf{S}=2\,\frac{\partial \Psi}{\partial \mathbf{C}},
  \qquad
  \mathbf{S}=\mathbf{F}^{-1}\mathbf{P},
\end{equation}
where $\mathbf{P}$ is the first and $\mathbf{S}$ the second Piola–Kirchhoff stress tensor. $\mathbf{C}$, $\mathbf{E}$, and $\mathbf{S}$ are symmetric tensors.
We employ a simple hyperelastic material model in this work, following a compressible neo-Hookean formulation.
With Lam\'e parameters $\mu$ and $\lambda$, its energy in 2D is defined as:
\begin{equation}\label{eq:neo_hooke_energy}
  \Psi(\mathbf{F})=\frac{\mu}{2}\,(\mathrm{tr}(\mathbf{C})-2)\;-\;\mu\,\ln J\;+\;\frac{\lambda}{2}\,(\ln J)^2.
\end{equation}
The second Piola-Kirchhoff stress then follows as
\begin{equation}
  \mathbf{S}=\mu\,(\mathbf{I}-{\mathbf{C}}^{-1})\;+\;\lambda\,\ln J\;{\mathbf{C}}^{-1}.
\end{equation}

In a two-scale setting (\fetwo{}), the macroscopic problem on domain $\Omega$ provides the macroscopic deformation gradient $\mathbf{F}^{\Omega}$ to an RVE that resolves equilibrium in the microscale domain $\omega$.
The microscale problem returns the homogenized stress as a volume average:
\begin{equation}
    \mathbf{P}^{\Omega} = \dfrac{1}{|\omega|} \int_{\omega} \mathbf{P}^{\omega}\, d\omega.
\end{equation}

\section{Microstructure-conditioned PRNNs}
\subsection{PRNNs for finite strain}
In this work we will use PRNNs as a surrogate for the microscale model, building on earlier work that demonstrated that PRNNs perform well in a finite strain framework \citep{MAMaia2024}.
The PRNN consists in an \textit{encoder - material - decoder} framework, with operations that conceptually mirror computational homogenization in an RVE.
The encoder learns to compute microscopic strain quantities from the macroscopic strain input.
These microscopic strains are passed through a set of \textit{fictitious material points} employing physics-based microscopic material models to produce microscopic stresses.
These material models are the same ones also used as constitutive equations in the RVE for data generation, in this case neo-Hookean.
The decoder then combines the microscopic stresses into a macroscopic stress.
The general flow of the PRNN used in this work is thus as follows:

\begin{equation}
    \mathbf{F}^{\Omega} \;\rightarrow\; \mathbf{E}^{\Omega}
    \underbrace{\;\rightarrow\; \hat{\mathbf{E}}^{\omega}_{m_1,\dots,M} }_{\text{encoder}}
    \underbrace{\;\rightarrow\; \hat{\mathbf{S}}^{\omega}_{m_1,\dots,M}}_{\text{material model}}
    \underbrace{\;\rightarrow\; \hat{\mathbf{S}}^{\Omega} \vphantom{\hat{\mathbf{C}}^{\omega}_{m_1,\dots,M}}}_{\text{decoder}}
    \;\rightarrow\;
    \hat{\mathbf{P}}^{\Omega}.
\end{equation}
Here, $M$ is a hyperparameter that corresponds to the number of fictitious material points used.
Note that these microscopic quantities are not linked to specific points in the micromodel.
Depending on the architecture, the distribution of the fictitious internal variables can still closely match those of the RVE \citep{Maia2026}.
However, unlike in the RVE, compatibility and equilibrium of these microscopic quantities are not enforced.

In the small-strain plasticity setting that inspired the original PRNN \citep{Maia2023}, most of the nonlinearity comes from the material behavior, which is captured by the material model present in the network.
In our finite-strain hyperelastic model, a significant amount of additional geometric nonlinearity can occur.
With the material model capturing material nonlinearity, and the decoder conceptually mirroring the homogenization step, the encoder is the natural place for capturing geometric effects.
We therefore explore a nonlinear encoder with a high flexibility.
In addition, PRNNs can be made thermodynamically consistent \citep{Maia2026}.
This works when summing the strain energy densities resulting from the material models, or equivalently, fixing the decoder to be a scaled transpose of the encoder.
We use this energy based PRNN as alternative to the nonlinear PRNN.

There are two conditions we want the PRNNs to satisfy.
First, in the undeformed state, all microscopic strains should vanish: $\mathbf{F}^{\Omega}=\mathbf{I} \to \hat{\mathbf{E}}^{\omega}_m = \mathbf{0}$.
This causes all microscopic stresses and energies to remain zero, leading to a zero predicted stress.
Second, to be able to compute the response of the fictitious material points, which are given by a classical neo-Hookean model, the encoder must output valid deformation tensors.
For this neo-Hookean model, this means ensuring that $\hat{\mathbf{C}}^{\omega}_m$ is positive definite.

\subsubsection{Nonlinear PRNN}
The first formulation we explore is designed around enabling a high flexibility in the model.
The approach we take is to first allow arbitrary nonlinear operations on the inputs, and then constrain this intermediate state to satisfy the two conditions outlined previously.

For the encoder, we use a neural network, $\mathrm{NN}^E$ that takes $\mathbf{E}^{\Omega}$ as input and outputs $3M$ variables (three local strain components per fictitious material point), gathered in $\boldsymbol{\theta}$.
Considering the first condition, we apply a zero-shift to ensure $\boldsymbol{\theta}$ is zero in the undeformed state:
\begin{equation}\label{eq:zero-shift}
    \bar{\boldsymbol{\theta}} = \mathrm{NN}^E(\mathbf{E}^{\Omega}) - \mathrm{NN}^E(\boldsymbol{0}).
\end{equation}
Each $\bar{\boldsymbol{\theta}}_m \in \mathbb{R}^3$ is then mapped to $\hat{\mathbf{C}}^{\omega}_m$ through a Cholesky-style parametrization,
\begin{equation}
    \hat{\mathbf{C}}^{\omega}_m = \mathbf{L}_m \mathbf{L}_m^{\top},
    \qquad \mathbf{L}_m = \begin{bmatrix} \sigma(\theta_{m,1}) & 0 \\ \theta_{m,2} & \sigma(\theta_{m,3}) \end{bmatrix},
\end{equation}
where $\sigma(x) = \mathrm{softplus}(x) - \ln(2) + 1$ is a shifted softplus chosen so that $\sigma(0)=1$.
This ensures that when $\bar{\boldsymbol{\theta}}_m$ is zero, $\hat{\mathbf{C}}^{\omega}_m=\mathbf{I}$, satisfying the first condition.
In addition, the resulting $\hat{\mathbf{C}}^{\omega}_m$ is positive definite by definition, and thus satisfies the second condition.

\begin{algorithm}[H]
\caption{Nonlinear encoder layer in 2D}
\label{alg:nonlinear_encoder}
\begin{algorithmic}[1]
    \Require $\mathbf{E}^{\Omega} \in \mathbb{R}^{2 \times 2}$, neural network $\mathrm{NN}^E$ with output size $3M$
    \Ensure $\hat{\mathbf{C}}^{\omega}_m \in \mathbb{R}^{2 \times 2}$ for $m = 1, \dots, M$
    \State Evaluate network and apply zero-shift: $\bar{\boldsymbol{\theta}} = \mathrm{NN}^E(\mathbf{E}^{\Omega}) - \mathrm{NN}^E(\mathbf{0})$
    \For{$m = 1, \dots, M$}
        \State Construct lower triangular matrix:
      $\mathbf{L}_m = \begin{pmatrix} \operatorname{softplus}(\theta_{m,1}) - \ln(2) + 1 & 0 \\ \theta_{m,2} & \operatorname{softplus}(\theta_{m,3}) - \ln(2) + 1 \end{pmatrix}$
        \State $\hat{\mathbf{C}}^{\omega}_m = \mathbf{L}_m \, \mathbf{L}_m^\top$

    \EndFor
\end{algorithmic}
\end{algorithm}

$\hat{\mathbf{C}}^{\omega}_m=$ is passed through the material model to obtain $\hat{\mathbf{S}}^{\omega}_{m}$.
Then a decoder combines all predicted microscopic stress tensors into the macroscopic stress.
Previous work on PRNNs for finite strains compared various decoders \citep{MAMaia2024}.
We adopt a sparse decoder, which combines the stresses in a component-wise manner:
\begin{equation}
    \hat{\mathbf{S}}^{\Omega}_i = \sum_m \hat{\mathbf{S}}^{\omega}_{m,i} \operatorname{softplus}(w_{m, i}).
\end{equation}

\subsubsection{Energy PRNN}

As an alternative to the nonlinear PRNN, we follow earlier work on creating thermodynamically consistent PRNNs \citep{Maia2026} and present an energy-based model.
The general idea is to compute the strain energy density as a weighted sum of the strain energy densities of the fictitious material points:

\begin{equation}\label{eq:energy_decoder}
\Psi^{\Omega} = \sum_m |w_m| \hat{\psi}^{\omega}_m
\end{equation}
with $w_m$ a scalar learned by the PRNN.

To satisfy the two conditions discussed earlier, we formulate the encoder based on the true strain $\mathbf{H}^{\Omega} = \tfrac{1}{2}\ln\bigl(\mathbf{C}^{\Omega}\bigr)$.
Here, the matrix logarithm $\ln(\mathbf{C}^{\Omega})$ is obtained by applying the logarithm to the eigenvalues of $\mathbf{C}^{\Omega}$ while retaining its eigenvectors.\footnote{Gradients of the eigendecomposition are singular for repeated eigenvalues (e.g., when $\mathbf{C}^{\Omega}=\mathbf{I}$). We therefore evaluate the logarithm for the symmetric positive definite $2\times2$ matrix with a closed-form expression that avoids explicit eigenvectors.}
The encoder is defined as:
\begin{equation}
    \label{eq:energy-encoder}
    \hat{\mathbf{C}}^{\omega}_m = \exp\bigl(2\hat{\mathbf{H}}^{\omega}_m\bigr),
    \qquad
    \hat{\mathbf{H}}^{\omega}_m = \mathbb{A}_m : \mathbf{H}^{\Omega},
\end{equation}
where $\mathbb{A}_m$ is a learned fourth-order tensor that applies a linear mapping for each fictitious material point.
The encoder thus maps strain linearly in log space.
The matrix exponential ensures that if $\hat{\mathbf{H}}^{\omega}_m$ is symmetric, then $\hat{\mathbf{C}}^{\omega}_m$ is symmetric positive definite.
Given that $\mathbf{H}^{\Omega}$ is symmetric, we can reduce $\mathbb{A}_m$ to have nine independent components and make $\hat{\mathbf{H}}^{\omega}_m$ symmetric by construction.
In Mandel notation $\{\mathbf{H}\} = [H_{11},\, H_{22},\, \sqrt{2}H_{12}]^{\top}$ we can write this as $\mathbf{A}_m \in \mathbb{R}^{3\times3}$:
\begin{equation}
    \label{eq:enc-mandel}
    \{\hat{\mathbf{H}}^{\omega}_m\} = \mathbf{A}_m\,\{\mathbf{H}^{\Omega}\},
    \qquad
    \mathbf{A}_m = \mathbf{I} + \operatorname{reshape}_{3\times3}\bigl(\boldsymbol{\theta}_m\bigr),
    \quad \boldsymbol{\theta}_m \in \mathbb{R}^9.
\end{equation}
In the undeformed state $\mathbf{H}^{\Omega} = \mathbf{0}$, the linear transformation results in $\hat{\mathbf{H}}^{\omega}_m = \mathbf{0}$ and $\hat{\mathbf{C}}^{\omega}_m = \mathbf{I}$.
For any admissible deformation, $\hat{\mathbf{C}}^{\omega}_m$ is symmetric positive definite.
Therefore, both conditions hold for every $\mathbf{A}_m$.
The identity in $\mathbf{A}_m$ is not strictly required, but it lets $\boldsymbol{\theta}_m$ represent only the deviation of $\hat{\mathbf{H}}^{\omega}_m$ from $\mathbf{H}^{\Omega}$.
If $\boldsymbol{\theta}_m$ is initialized close to zero, the model starts from the macroscopic response, which simplifies training.

The fictitious material points use $\hat{\mathbf{C}}^{\omega}_m$ and output a strain energy density $\hat\psi^\omega_m$.
The decoder, given in Equation \ref{eq:energy_decoder}, combines the strain energy densities.
The energy PRNN thus has 10 learnable parameters per material point.
The stress is obtained by differentiating $\Psi$, either using automatic differentiation or analytically using the chain rule.
As the stress is the exact gradient of a scalar potential, the tangent is symmetric with hyperelastic materials points and the network is thermodynamically consistent.
However, polyconvexity and rank-one convexity are not guaranteed, and any symmetry in the RVE is not enforced.

\begin{algorithm}[H]
\caption{Energy PRNN with a neo-Hookean material model in 2D}
\label{alg:hencky_layer}
\begin{algorithmic}[1]
  \Require $\mathbf{H}^{\Omega} \in \mathbb{R}^{2 \times 2}$, parameters $\boldsymbol{\theta} \in \mathbb{R}^{9 \times m}$, $\mathbf{w} \in \mathbb{R}^m$
  \Ensure Macroscopic stress $\hat{\mathbf{S}}^{\Omega}$
  \State $\hat{\Psi}^{\Omega} = 0$
  \For{$m = 1, \dots, M$}
      \State $\mathbf{A}_m = \mathbf{I} + \operatorname{reshape}_{3\times3}(\boldsymbol{\theta}_m)$
      \State $\{\hat{\mathbf{H}}^{\omega}_m\} = \mathbf{A}_m \{\mathbf{H}^{\Omega}\}$
      \Comment{Mandel notation, \eqref{eq:enc-mandel}}
      \State $\hat{\mathbf{C}}^{\omega}_m = \exp\bigl(2\hat{\mathbf{H}}^{\omega}_m\bigr)$
      \State $\ln \hat{J}^{\omega}_m = \operatorname{tr}\hat{\mathbf{H}}^{\omega}_m$
      \State $\hat{\psi}^{\omega}_m = \tfrac{\mu_m}{2}\bigl(\operatorname{tr}\hat{\mathbf{C}}^{\omega}_m - 2\bigr) - \mu_m \ln \hat{J}^{\omega}_m + \tfrac{\lambda_m}{2}\bigl(\ln \hat{J}^{\omega}_m\bigr)^2$
      \Comment{Specific to a neo-Hookean model.}
      \State $\hat{\Psi}^{\Omega} = \hat{\Psi}^{\Omega} + |w_m| \hat{\psi}^{\omega}_m$
  \EndFor
  \State $\hat{\mathbf{S}}^{\Omega} = \partial \hat{\Psi}^{\Omega} / \partial \mathbf{E}^{\Omega}$
\end{algorithmic}
\end{algorithm}

\subsection{A hypernetwork-based PRNN}\label{sec:hyprnn}
PRNNs cannot generally be conditioned on microstructural parameters apart from material parameters.
Different routes can be explored to still include them in the PRNN architecture.
The straightforward approach to condition a data-driven model on additional parameters is by concatenating the variables as additional inputs.
For the PRNN, this approach conflicts with the role of the encoder, which is meant to map only the macroscopic strain onto a latent representation of microscopic strains.
While the nonlinear encoder could in theory support additional inputs, the energy-based formulation would require further modifications.

\begin{figure}[htbp]
    \centering
    \includegraphics[width=1.\textwidth]{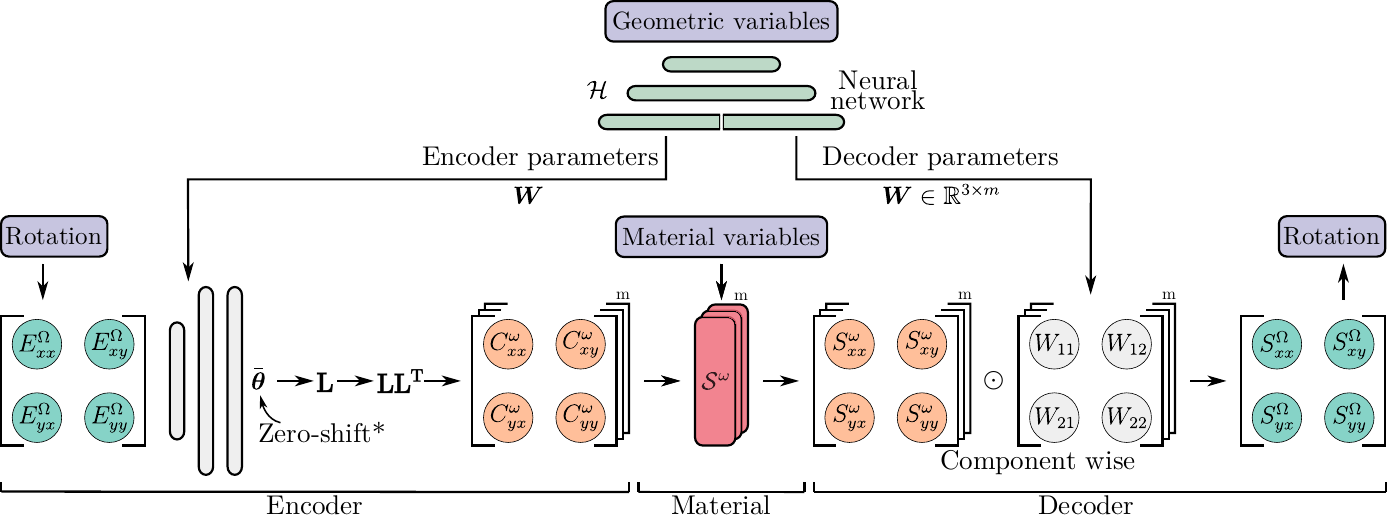}
    \caption{Overview of the nonlinear HyPRNN. *Zero-shift: see Equation \ref{eq:zero-shift}. 
    }
    \label{fig:HyPRNN_nonlinear}
\end{figure}

\begin{figure}[htbp]
    \centering
    \includegraphics[width=1.\textwidth]{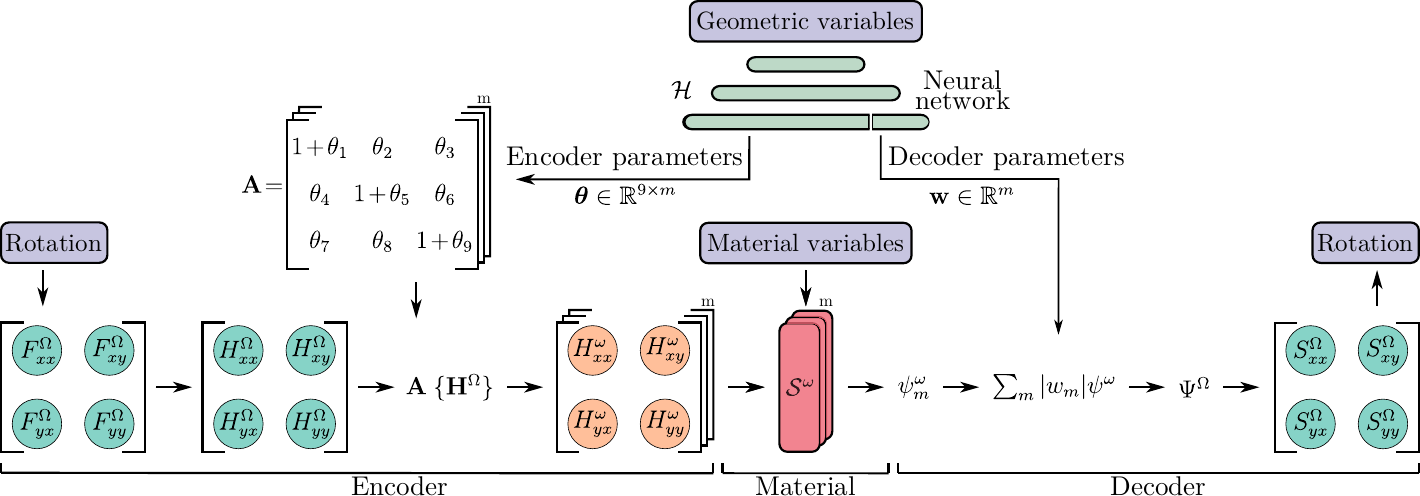}
    \caption{Overview of the energy HyPRNN. $\{ \cdot \}$ refers to a tensor in Mandel notation.
    }
    \label{fig:HyPRNN_energy}
\end{figure}

Instead, we preserve the existing PRNN architecture, and use a hypernetwork $\mathcal{H}$ that learns the PRNN encoder and decoder parameters based on microstructural variables.
In effect, the hypernetwork learns a continuous mapping from microstructural variables to the space of PRNN weights, rather than producing a single set of weights tied to one microstructure.
The encoder and decoder weights are learned by the same hypernetwork, and the only trainable parameters then become the hypernetwork parameters themselves.
These are learned in a single training stage, directly from RVE simulation data on different microstructures.
We provide a schematic overview of a nonlinear HyPRNN in Figure \ref{fig:HyPRNN_nonlinear}, and an energy HyPRNN in Figure \ref{fig:HyPRNN_energy}.

Geometric parameters of the microscale are used as inputs to $\mathcal{H}$.
Material parameters that affect the microscopic constitutive model can be directly modified in the fictitious material points on a per-sample basis.
This has been shown to work even for large changes of parameters, and no retraining of the network is required \citep{Kovcs2025}.
Microscale orientations can be accounted for by rotating the input deformation and output stress accordingly.

\section{Results: conditioning on microscale variables}\label{sec:res1}
In this results section, we compare a HyPRNN against a neural network and numerically validate it in a multiscale simulation, before performing a graded multiscale optimization.

\subsection{Numerical setup}\label{sec:setup}
This section describes the software and modeling choices used for the numerical experiments.
All code, as well as some pretrained surrogate models, is accessible on GitHub at \url{https://github.com/SLIMM-Lab/hyprnn}.

\subsubsection{Parametrization and data generation}\label{sec:param_and_data}
To train the surrogate models, a dataset is generated by simulating RVEs spanning a range of microstructural parameters.
The RVEs consist of random packings of ellipses, representing the woodchips, surrounded by the mycelium.
We consider four parameters: the ellipse aspect ratio $r$, the woodchip volume fraction $V_f$, the mycelium shear modulus $\mu$, and the ellipse orientation $\phi$.
The first two are geometric parameters, and $\mu$ is a material parameter
As all ellipses follow the same orientation $\phi$, it can be accounted for by rotating the deformation to align with the ellipse direction before passing it to the PRNN.
Since we limit ourselves to 2D in this work, this is computed as:
\begin{equation}
\bar{\mathbf{F}} = \mathbf{Q}^T \mathbf{F} \mathbf{Q}, \quad \mathbf{Q} =\begin{pmatrix} \cos\phi & -\sin\phi \\ \sin\phi & \cos\phi \end{pmatrix}.
\end{equation}
The output is then back-rotated to obtain the actual stress $\mathbf{S}$.
\begin{equation}
\mathbf{S} = \mathbf{Q} \bar{\mathbf{S}} \mathbf{Q}^T
\end{equation}
This way the orientation can be included by pre- and post-processing the surrogate's inputs and outputs, and does not need to be included in the dataset.

The material variable $\mu$ is included in the dataset, but it is not necessary for the random ellipse packing of the RVE.
This packing is parametrized based on $r$ and $V_f$.
$V_f$ determines the number of ellipses, and $r$ is are shared for all ellipses within one RVE.
For a given aspect ratio, we compute the length and width of the ellipses such that the area of each ellipse remains constant, covering 0.8\% of the RVE domain.
Ellipses are then iteratively placed at random coordinates in a square domain until the desired volume fraction is reached.
Ellipses crossing a boundary are wrapped to the opposite boundary to maintain a periodic geometry.
If a newly placed ellipse overlaps with an existing one, it is discarded and a new random location is attempted.
Gmsh \citep{Geuzaine2009} is used to mesh the domain.
To produce a robust mesh, we verify that the number of nodes on opposite edges matches, and restart the procedure if this is not the case.

We simulate the RVEs using \textit{DOLFINx} \citep{baratta2023dolfinx} with periodic boundary conditions.
The load direction is selected randomly in $\mathbf{F}$ space, and the RVE is loaded monotonically for 50 time steps up to $ \|\mathbf{F} - \mathbf{I}\| = 0.5$.
If a time step fails to converge, the state is reverted to the last converged step, and the step is repeated with eight sub-increments.
However, data from the intermediate sub-increments is not included in the dataset.
If sub-increments also fail to converge, the simulation is stopped and only the steps up to the last converged step are included in the dataset.

\begin{figure}[htbp]
    \centering
    \includegraphics[width=1.0\textwidth]{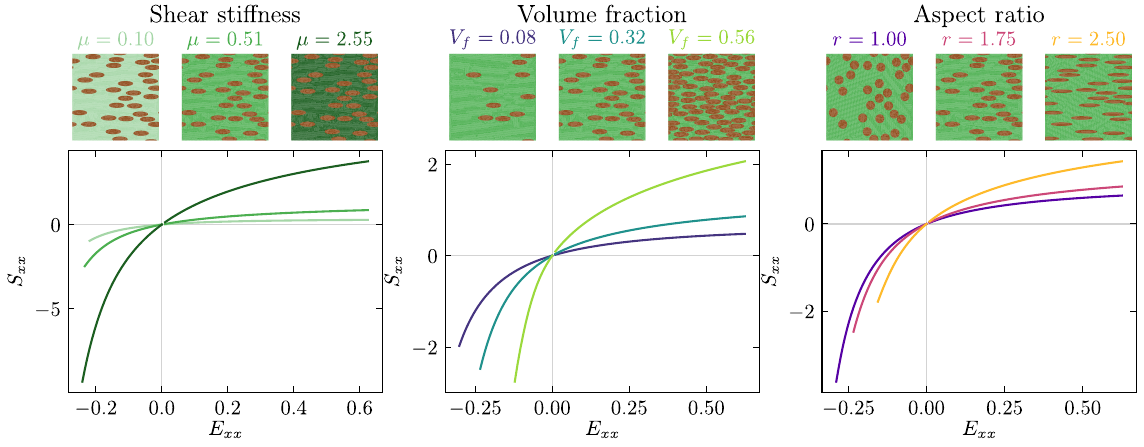}
    \caption{Influence of the microstructural parameters. $\mu=0.51$, $V_f=0.32$, $r=1.75$ are used as default values; one of the variables is set to the minimum or maximum value in the dataset. The sample is displaced in the $E_{xx}$ component only, in both tension and compression.}
    \label{fig:data_variation}
\end{figure}

The influence of changing the microstructural parameters on the stress response is shown in Figure \ref{fig:data_variation}.
In these responses, only the $E_{xx}$ component is loaded for illustrative purposes.
In tension, the simulations succeed for all 50 time steps.
However, in compression, the simulations fail significantly earlier.
Similar stress differences to those in this figure also occur for the other stress directions.

\subsubsection{Surrogate model training}
The HyPRNN, including the embedded material model, is implemented in JAX \citep{Bradbury2018}.
Two-thirds of the dataset is used for training, and the remainder is split evenly between validation and test sets.
The training is done in $\mathbf{E}$-$\mathbf{S}$ space, since their symmetry allows for reducing the input and output dimensions of the surrogates, which is generally advantageous for data-driven models.
An L2 loss is minimized using the Adam optimizer \citep{Kingma2014}, with a fixed learning rate of $0.001$, until the validation loss has not improved for 50 epochs.
All computations, including the training of the surrogates, are performed on a CPU.
Just-in-time compilation and vectorized computations enable efficient training and inference.
Since there is no history dependency considered here, all steps along a load path are computed in a single batched operation, together with steps from other load paths.
A batch size of 2 load paths is used for all models.

\subsubsection{Multiscale simulations}
All macroscopic simulations are also performed in \textit{DOLFINx}, where the \textit{dolfinx\_materials} \citep{Bleyer2024} library enables the use of RVE micromodels and JAX-based surrogates as constitutive models.
For surrogate-based multiscale simulations, automatic differentiation is used to compute the tangent stiffness tensor.
Just-in-time compilation is used here too, resulting in a longer computation time for the first time step, but then enabling much faster subsequent steps.
For the RVE-based multiscale simulations, we compute the tangents with central difference for ease of implementation.
The RVE is solved with a solver tolerance of $10^{-10}$, the central difference perturbation is $10^{-8}$. The macroscale problem is solved with a tolerance of $10^{-6}$.
Substepping with eight sub-increments is applied at the macroscale in case of non-convergence.
The micromodels are evaluated in parallel.

\subsection{Conditioning comparisons}
In this section, we compare the different HyPRNN architectures.
We additionally compare their performance to that of a standard neural network.
Since PRNNs with a linear encoder can extrapolate material parameters after having been trained with only one \citep{Kovcs2025}, we first separately consider the effect of conditioning only on material parameters or only on geometric parameters, before combining them.

Three datasets, $\mathcal{D}^{mat}$, $\mathcal{D}^{geo}$, and $\mathcal{D}^{all}$, are generated to test the influence of different microstructural parameters.
For $\mathcal{D}^{mat}$, only $\mu$ varies, for $\mathcal{D}^{geo}$, $V_f$ and $r$ change, and for $\mathcal{D}^{all}$ all three parameters are varied.
An overview is given in Table \ref{tab:datasets}.
The shear modulus $\mu$ is selected randomly in the range $[0.10, 2.55]$.
Since whole ellipses are placed one at a time, $V_f$ is restricted to discrete increments.
Rather than generating a new mesh for every $(V_f,r)$ combination that would arise from uniform sampling of $r$, we draw from a predefined set of $(V_f,r)$ meshes.
The combinations are a grid with $V_f \in \{0.08, 0.16, 0.24, 0.32, 0.40, 0.48, 0.56\}$, and $r \in \{1.00, 1.25, 1.50, 1.75, 2.00, 2.25, 2.50\}$, giving a total of 49 combinations.

\begin{table}[htbp]
  \centering
  \caption{Overview of the three datasets. Fixed parameters show their value; varying parameters show their range.}
  \label{tab:datasets}
  \begin{tabular}{l ccc ccc}
    \toprule
    & \multicolumn{3}{c}{Parameters} & \multicolumn{3}{c}{\# Samples} \\
    \cmidrule(lr){2-4} \cmidrule(lr){5-7}
    & $\mu$ & $V_f$ & $r$ & Train & Val & Test \\
    \midrule
    $\mathcal{D}^{mat}$ & $[0.10,\; 2.55]$ & $0.40$ & $2.00$ & 2048 & 512 & 512 \\
    $\mathcal{D}^{geo}$ & $0.51$ & $\{0.08, \ldots, 0.56\}$ & $\{1.00, \ldots, 2.50\}$ & 2048 & 512 & 512 \\
    $\mathcal{D}^{all}$ & $[0.10,\; 2.55]$ & $\{0.08, \ldots, 0.56\}$ & $\{1.00, \ldots, 2.50\}$ & 2048 & 512 & 512 \\
    \bottomrule
  \end{tabular}
\end{table}

We create learning curves for different hyperparameter settings of the neural network, the nonlinear HyPRNN, and the energy HyPRNN.
For $\mathcal{D}^{mat}$, no hypernetwork is necessary in the PRNN as $\mu$ directly affects the fictitious material points. 
The ability of a PRNN to predict for different material parameters was already shown for plasticity in \citep{Kovcs2025}.
In contrast to that study, where only a single value is seen in training, the material parameter is varied in the training data here, in order to allow for a meaningful comparison with a neural network.
The neural network, like the nonlinear PRNN, predicts $\mathbf{S}^{\Omega}$ from $\mathbf{E}^{\Omega}$, with microscopic variables as additional inputs.

\begin{figure}[htbp]
    \centering
    \begin{subfigure}[b]{0.325\textwidth}
        \centering
        \includegraphics[width=\textwidth]{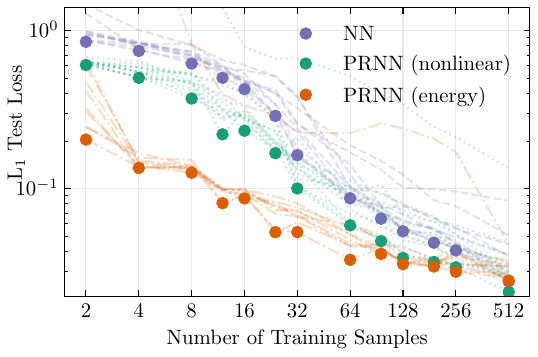}
        \caption{$\mathcal{D}^{mat}$}
        \label{fig:learn_curve_mu}
    \end{subfigure}
    \hfill
    \begin{subfigure}[b]{0.325\textwidth}
        \centering
        \includegraphics[width=\textwidth]{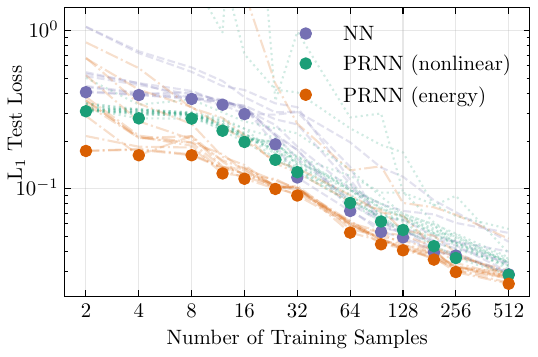}
        \caption{$\mathcal{D}^{geo}$}
        \label{fig:learn_curve_vfrac_ratio}
    \end{subfigure}
    \hfill
    \begin{subfigure}[b]{0.325\textwidth}
        \centering
        \includegraphics[width=\textwidth]{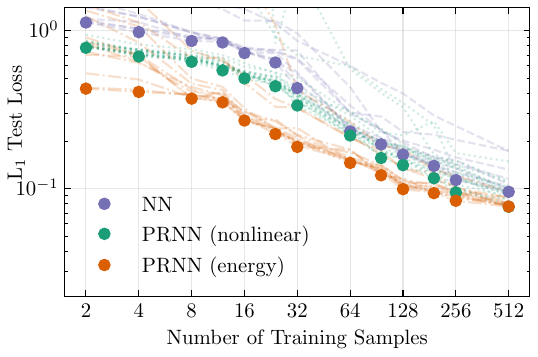}
        \caption{$\mathcal{D}^{all}$}
        \label{fig:learn_curve_all}
    \end{subfigure}
    \caption{Overview of learning curves on the different datasets, plotted on a log-log scale. Each line represents the learning curve for one configuration of hyperparameters, averaged over 10 runs. The lowest point for each type of model (NN, nonlinear (Hy)PRNN, energy (Hy)PRNN) is highlighted to show the overall trend.}
    \label{fig:learn_curves}
\end{figure}

The hyperparameters include the choice of the activation functions and the network size, which for the PRNN corresponds to the number of material points.
A more detailed overview of hyperparameters considered is given in \ref{app:hyperparameters}.
Each configuration is trained 10 times per number of training samples with different samples drawn from the training pool, and the average loss over these 10 training runs is plotted in Figure \ref{fig:learn_curve_mu}.
Each plot includes learning curves across a wide range of hyperparameters, and therefore different levels of complexity.
This avoids having to either separately perform model selection for each dataset size, or first perform model selection on the largest dataset size and then risk biasing our results on smaller dataset sizes due to overfitting.
The lowest average loss per type is marked with a circle, giving an envelope of the best models.
All models improve consistently as more training data is added.
The nonlinear PRNN achieves lower losses than the NN.
The energy PRNN performs better still, particularly in the low-data regime, where it reaches a low loss with only a few samples.
As the number of samples increases, the nonlinear PRNN and NN losses catch up with the energy PRNN.
Still, it is remarkable how the energy PRNN can achieve consistently high accuracies here even though --- in contrast to the nonlinear encoder one --- it has no internal mechanism to capture geometric nonlinearities.

The same process is repeated for $\mathcal{D}^{geo}$, and the results are shown in Figure \ref{fig:learn_curve_vfrac_ratio}.
Here, the conditioning happens solely via the hypernetwork.
A similar trend to before is observed, with the energy HyPRNN performing better in the low-data regime, but all models obtaining similar losses for large sample sizes.
The error decreasing to a low value indicates that the hypernetwork approach to conditioning a PRNN is effective.

Figure \ref{fig:learn_curve_all} shows the resulting learning curves for $\mathcal{D}^{all}$, combining both the geometric and material parameters.
The first thing that can be observed is that the error is higher than in the other two learning curve plots.
This is because the dataset is more complex, and also reaches higher stress values, making the absolute difference larger than the relative difference in prediction.
Due to the definition of the range of $\mu$, the average $\mu$ in $\mathcal{D}^{all}$ is also higher than $\mu$ in $\mathcal{D}^{geo}$.
The gap in Figure \ref{fig:learn_curve_mu} between the energy PRNN and the NN in the low-data regime is larger than the gap in Figure \ref{fig:learn_curve_vfrac_ratio}, confirming the particular benefit that the PRNN with a linear transformation in the encoder offers in that it can transfer between different material property values without being trained for it.
The nonlinear PRNN loss is closer to that of the NN, showing that it does not transfer material properties as well as an energy PRNN.
This is because the nonlinear PRNN relies less strongly on the physics in the material model for reproducing the data.

For all the datasets, the differences between the types of models are significant, and variations for specific hyperparameters within one type are smaller.
This indicates that the overall results are not overly sensitive to the specific settings used, such as the number of material points or a specific number of neurons in a layer.
Sigmoid activation performs much better than ReLU activation in all networks (NN, nonlinear encoder, hypernetwork).
Furthermore, HyPRNN models with only the mycelium material points perform better than those with additional woodchip material points when there is little available data.
These effects on the learning curves are shown in \ref{app:hyperparameters}.
We suspect that the large contrast between the stiffness of the mycelium and the woodchip degrades training performance.
This phenomenon deserves more attention in future work.

\begin{figure}[htbp]
    \centering
    \includegraphics[width=1.\textwidth]{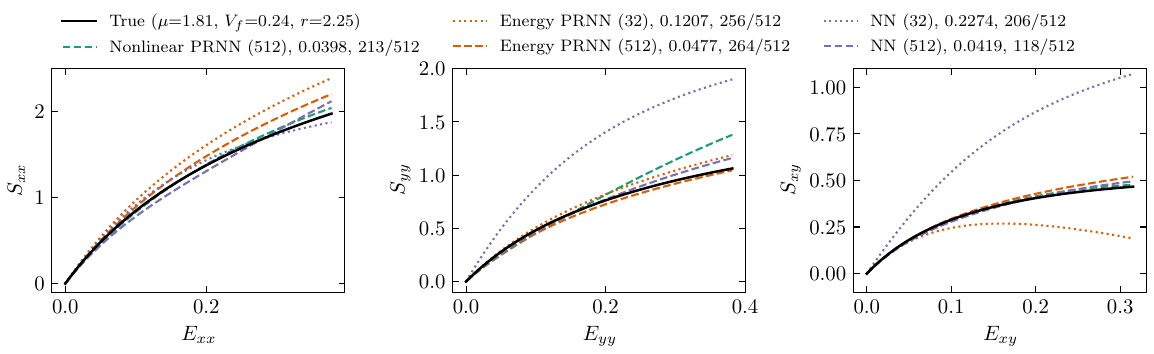}
    \caption{For specific model architectures, the instances with the median test losses are selected (out of 10 training runs) such that the responses represent that of a single training run. For these models, predictions are made on the sample for which the energy HyPRNN with 32 samples has the median error. The legend description is as follows: [model name (number of training samples), error on this sample, rank of sample error out of all test errors (1 is best, 512 is worst)]. The nonlinear HyPRNN and NN perform better on this sample than on their median sample, making the quality of their predictions in this plot lie on the optimistic side.}
    \label{fig:pred_result}
\end{figure}

A test curve from $\mathcal{D}^{all}$ with predictions from several representative models is visualized in Figure \ref{fig:pred_result} with pair-wise stress-strain plots.
For this specific curve, all surrogates predict the same overall shape, but quantitatively the nonlinear HyPRNN performs best.
The NN trained with 32 samples shows a significantly higher error compared to the other curves.
The energy PRNN predictions also give a relatively high error for this sample.
The nonlinear HyPRNN used here has twelve material points, a single hidden layer in the encoder, and three hidden layers in the hypernetwork, which is eight nodes wide and uses sigmoid activations.
The energy HyPRNN has twelve material points and a hypernetwork with a single hidden layer with sigmoid activation.
The neural network has four hidden layers of 64 nodes, with sigmoid activation.

\subsection{\fetwo{} validation}\label{sec:fe2_validation}
\begin{figure}[htbp]
    \centering
    \includegraphics[width=1.\textwidth]{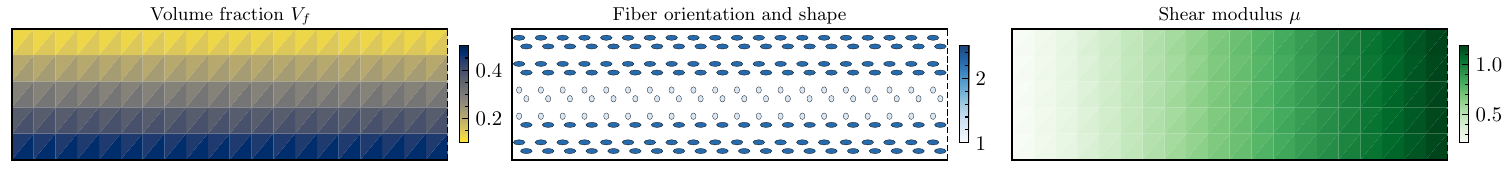}
    \caption{The applied grading throughout the domain of the 3-point bending beam. Only the left half of the domain is visualized. The volume fraction gradually increases from top to bottom. The woodchips are oriented horizontally and slender at the top and bottom, and more rounded and vertically oriented along the middle of the beam. The mycelium shear modulus is low near the supports, and increases gradually towards the center.}
    \label{fig:val_grading}
\end{figure}

Dataset losses quantify pointwise accuracy, but do not fully correlate with the surrogate performance in a full \fetwo{} simulation.
The nonlinear interactions between the points might cause the solution to drift, the tangent stiffness plays a direct role, and the distribution of explored points differs from the test set distribution.
We therefore validate our method on a simple structural example: a 3-point bending beam.
The beam measures $4.0 \times 0.3$ [m], and the top is vertically loaded in five increments over a width of $0.2$ [m] at the midspan.
Symmetry is used to simulate only the left half of the beam, fixing the bottom left corner, and preventing horizontal movement at the symmetric edge.
Note that this setup somewhat overconstrains a typical 3-point bending beam, as it does not allow for any horizontal movement.
The applied grading of the simulation is shown in Figure \ref{fig:val_grading}, spanning a wide range of the considered parameters.
The volume fraction gradually increases along the beam height; the shear modulus increases towards the beam center.
The woodchip aspect ratio and orientation, in contrast, jump discontinuously between an inner and an outer band along the height.

\begin{figure}[htbp]
    \centering
    \includegraphics[width=1.\textwidth]{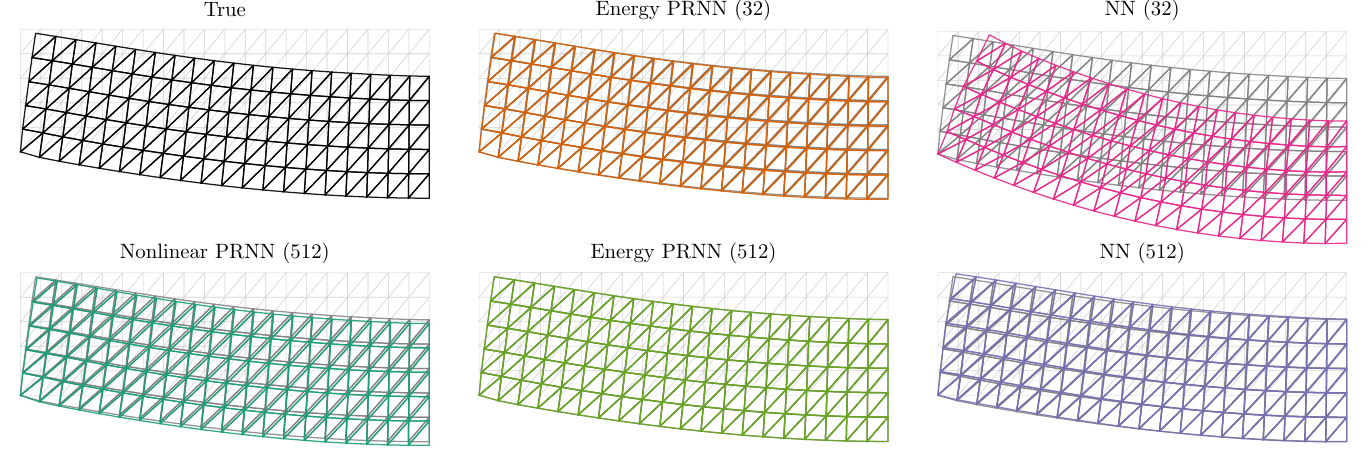}
    \caption{Deformations following the multiscale simulations of a 3-point bending beam. Each plot shows the initial mesh in light gray, the ground truth RVE-based \fetwo{} simulation in dark gray, and the surrogate-based simulation.}
    \label{fig:val_deformation}
\end{figure}

For the energy HyPRNN and the NN, two variations are considered, one with 32 training samples and one with 512 training samples.
For the nonlinear HyPRNN, we only consider a model trained on 512 samples.
A comparison of the displacement fields of surrogate-based simulations to the RVE-based simulation is visualized in Figure \ref{fig:val_deformation}.
In the RVE-based reference solution the midspan deflection is $0.115$ [m] with strains reaching 27\%.
All HyPRNN-based surrogates show relatively good agreement, and the energy HyPRNN with 512 samples shows a near-perfect match.
While the NN and nonlinear HyPRNN with 512 samples reach similarly low errors on the test set, their displacements in these multiscale simulations are less accurate than the energy HyPRNN.
The NN with 32 samples shows very poor performance.
In \ref{app:fe2val}, the stress strain paths at a few quadrature points are shown, clearly showing how the neural network surrogate produces a nonzero stress in the undeformed state, whereas the HyPRNN models do not.

We compare the computational cost of the different simulations.
All computation times are obtained on a standard desktop PC with an Intel Xeon W-2223 processor.
As discussed earlier, the RVE-based simulation uses central differences to compute the tangent, which is suboptimal in terms of the simulation time.
The results are presented in Table \ref{tab:fe2_time}, showing the much lower simulation time with the surrogate models.
The data generation and training times are one-time costs, whereas the online simulation time is repeated for each new graded simulation.
Still, this large data generation time illustrates the relevance of data requirements for surrogate modeling in multiscale analysis.
The first and remaining time steps are reported separately, since the first time step includes the just-in-time compilation of the surrogates.
The stark contrast between the RVE and surrogate based simulation time highlights the benefit of surrogate-based simulations.
This contrast would only increase for larger macroscopic simulations with more elements, or those running for more time steps.

\begin{table}[htbp]
  \centering
  \caption{Computation wall-time in seconds for multiscale simulations with the different constitutive models and their number of training samples indicated in brackets. For the RVE-based simulation, initialization includes creating the RVE meshes. Both the meshing and the RVE simulations are performed in parallel across 8 processes. The RVE simulation and data generation timings are from a single run; the surrogate training and simulation timings are averaged across three runs. Data-generation times are identical because they use the same data pool.}
  \label{tab:fe2_time}
  \begin{tabular}{ll ccccc}
    \toprule
    \multicolumn{2}{l}{Constitutive model}
    & RVE & Energy & Energy & Nonlinear & NN \\
    \multicolumn{2}{l}{(training samples)}
    &  & PRNN (32) & PRNN (512) & PRNN (512) & (512) \\
    \midrule
    \multirow{2}{*}{\shortstack[l]{Offline \\ stages}}
    & Data generation & -- & 2035.97 & 32575.51 & 32575.51 & 32575.51 \\
    & Training & -- & 53.98 & 230.86 & 1134.89 & 410.09 \\
    \midrule
    \multirow{4}{*}{\shortstack[l]{Online \\ simulations}}
    & Total & 5900.52 & 2.41 & 2.17 & 2.12 & 1.00 \\
    \cmidrule{2-7}
    & Initialization & 137.49 & 0.18 & 0.25 & 0.44 & 0.37  \\
    & Time step 1 & 965.75 & 2.16 & 1.85 & 1.56 & 0.59  \\
    & Time steps 2-5 & 4797.27 & 0.06 & 0.07 & 0.12 & 0.04 \\
    \bottomrule
  \end{tabular}
\end{table}

\subsection{Graded multiscale optimization}\label{sec:grad_opt}
\begin{figure}[htbp]
    \centering
    \includegraphics[width=.35\textwidth]{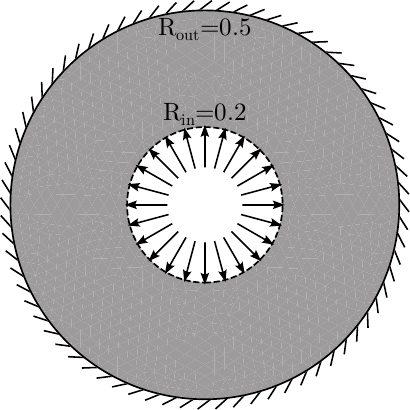}
    \caption{Problem setup for the disk loaded with internal pressure. A distributed load is applied to the inner boundary, while the outer boundary is fixed.}
    \label{fig:disk_overview}
\end{figure}

We now turn to using our surrogate as a constitutive model to optimize a multiscale graded material.
Disks subject to rotation or internal loads are commonly studied in the functionally graded materials literature \citep{Durodola2000,Bayat2008,Abdalla2020}, where a graded thickness is typically used to reduce peak stresses.
We adopt a similar setup but grade the microstructural parameters rather than the thickness under an internal pressure.
An overview of the structure and boundary conditions is provided in Figure~\ref{fig:disk_overview}.
The goal is to minimize peak stresses.
The parameters only vary radially, allowing for a 1D parametrization.
The aspect ratio and woodchip angle are combined into $r_{\mathrm{combi}}$ to enable rotated woodchips in a radially consistent manner.
Specifically, $r_{\mathrm{combi}} \in [0.4, 2.5]$, where values below one are equivalent to setting $r=1/r_{\mathrm{combi}}$ with angle $\phi=\ang{90}$ with respect to the radial direction, while for values above one, $r=r_{\mathrm{combi}}$ and $\phi=\ang{0}$.
In this parametrization, the woodchips can thus only be either parallel or perpendicular to the radial direction.
The optimization variables are $V_f$, $r_{\mathrm{combi}}$, and $\mu$, and we use the energy PRNN with 512 training samples.

The optimization problem is defined as minimizing the peak Cauchy stress under a given load, subject to a maximum acceptable deformation.
Accounting for symmetry, only one quarter of the disk is simulated.
The maximum acceptable deformation is set to 0.05 (16.6\% of the thickness) through a linearly scaling penalty on the loss when exceeded.
We use the Covariance Matrix Adaptation Evolution Strategy (CMA-ES) \citep{Hansen2001}, an off-the-shelf, iterative, derivative-free algorithm, to optimize our objective function.
Alternative optimization algorithms, such as Bayesian optimization, are also expected to work, and the choice of optimizer is not expected to affect our conclusions.

\begin{figure}[htbp]
    \centering
    \includegraphics[width=1.0\textwidth]{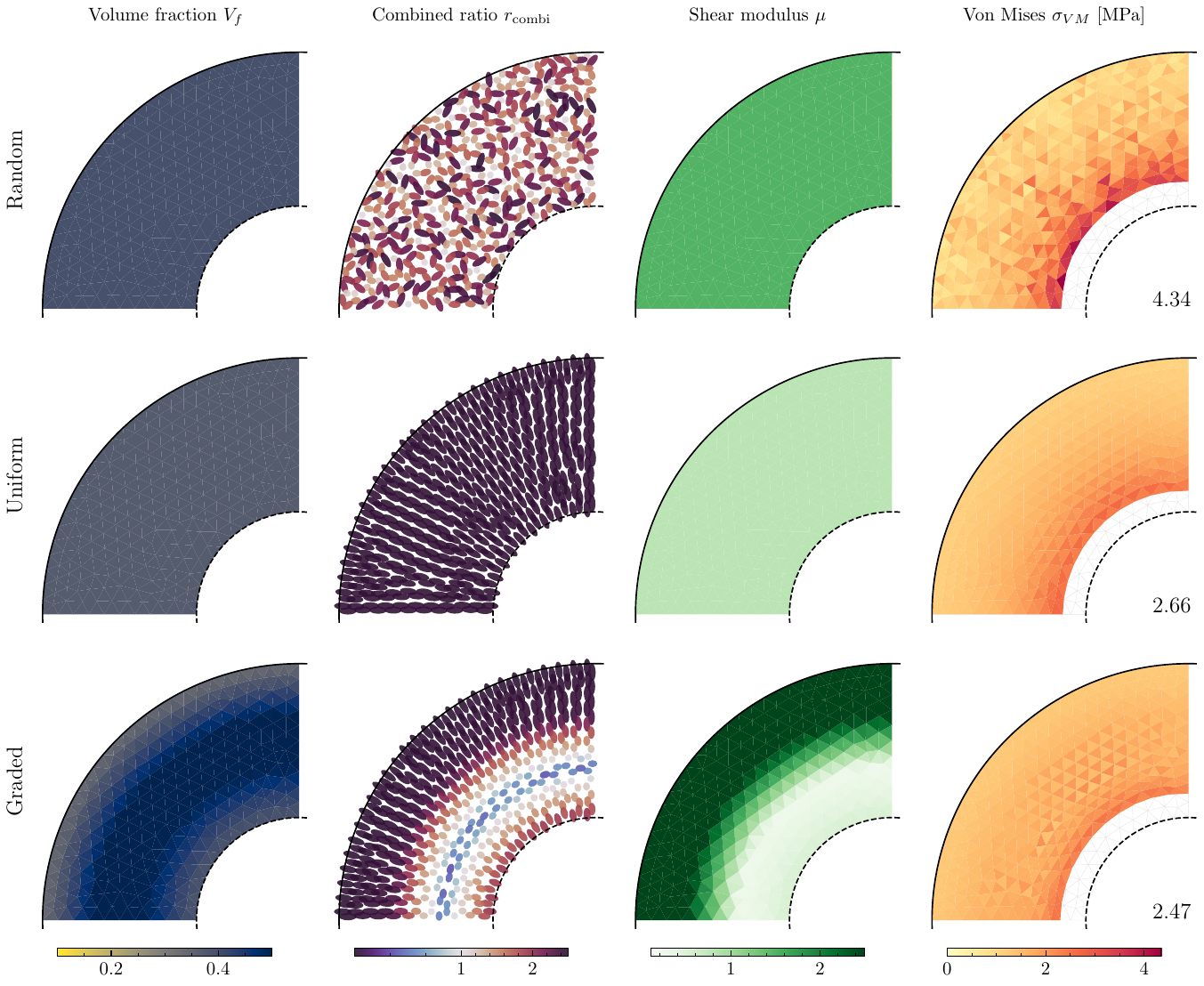}
    \caption{The optimized graded structures and their resulting average stresses. The average stresses are obtained by first computing the Cauchy stress, and then using the von Mises formula. The maximum stresses are 4.34, 2.66, and 2.47 [MPa] for the random, uniform, and graded structure respectively. The von Mises plot shows the stresses in the deformed configuration.}
    \label{fig:disk_grading_comparison}
\end{figure}

Three cases are compared: one random, one uniform, and one graded with four control points per variable through the thickness, between which the values are linearly interpolated.
The first of these is included as an unoptimized reference solution, while the other two represent optimization problems with increasing flexibility (three versus twelve parameters).
For the random initialization, the angle and aspect ratio are randomly chosen, $\mu$ is set to 1.5, and the volume fraction is modified to exactly satisfy the deformation constraint using the bisection method, leading to a peak stress of 4.34 [MPa].
The optimized results are visualized in Figure \ref{fig:disk_grading_comparison}.
An efficient uniform parametrization with a low peak stress of 2.66 [MPa] is reached after 1099 simulations.
The graded structure further decreases this peak stress to 2.47 [MPa] after 2163 simulations.
The optimization finds that setting $r_{\mathrm{combi}}$ below one is better inside the disk, rather than directly at the edge.

\section{Results: conditioning on manufacturing variables}\label{sec:res2}
Precisely controlling the microscale geometry as assumed in Section \ref{sec:res1} is not yet feasible for most materials.
Manufacturing process variables influence the microscale, but finding an exact relation between these variables and the microscale geometry is not straightforward.
We therefore showcase the ability to condition the PRNN on variables from the manufacturing process.
This serves as a computational proof-of-concept aiming to inspire future experimental work.

Mycelium-woodchip composites can be created by depositing woodchips from a bag or container into formwork, and then inoculating them with mycelium.
We simulate this deposition process and consider two variables in the process.
First, small round mycelium pellets can be added to the mix, decreasing the resulting woodchip volume fraction.
We assume that these have the same material properties as the mycelium that grows in the formwork.
Second, the initial orientation of the dropped woodchips can be controlled, for example by using a sieve, resulting in anisotropy.

\subsection{Dataset}
\begin{figure}[htbp]
    \centering
    \begin{subfigure}[b]{0.32\textwidth}
        \centering
        \includegraphics[width=\textwidth]{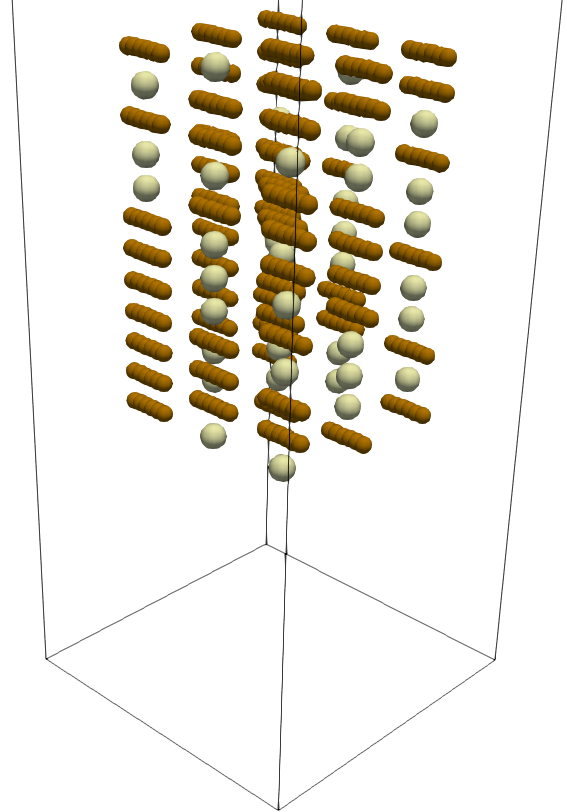}
        \caption{Initialization $\theta=$\ang{0}}
        \label{fig:grav_dep_3d_a}
    \end{subfigure}
    \hfill
    \begin{subfigure}[b]{0.32\textwidth}
    \centering
    \includegraphics[width=\textwidth]{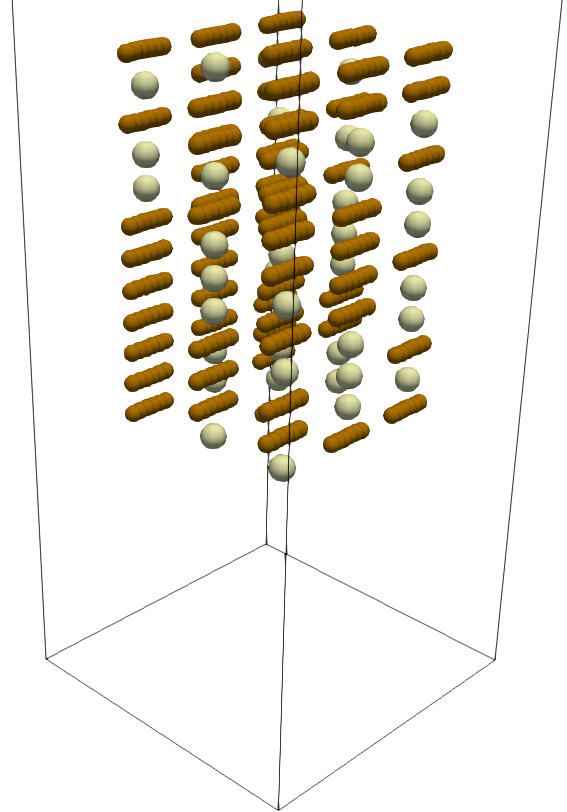}
    \caption{Initialization $\theta=$\ang{90}}
    \label{fig:grav_dep_3d_b}
    \end{subfigure}
    \hfill
    \begin{subfigure}[b]{0.34\textwidth}
        \centering
        \begin{subfigure}[b]{\textwidth}
            \centering
            \includegraphics[width=\textwidth]{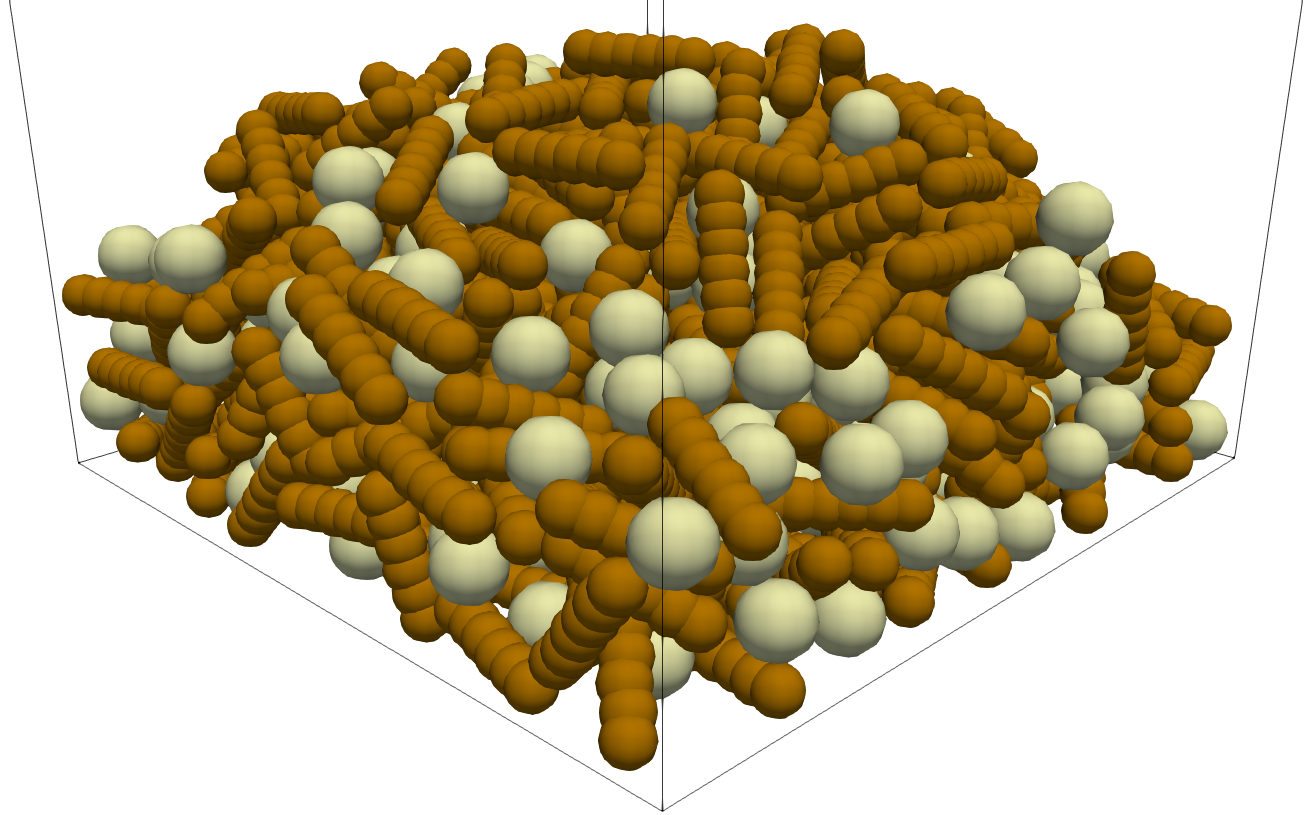}
            \caption{Settled deposition}
            \label{fig:grav_dep_3d_c}
        \end{subfigure}

        \begin{subfigure}[b]{\textwidth}
            \centering
            \includegraphics[width=\textwidth]{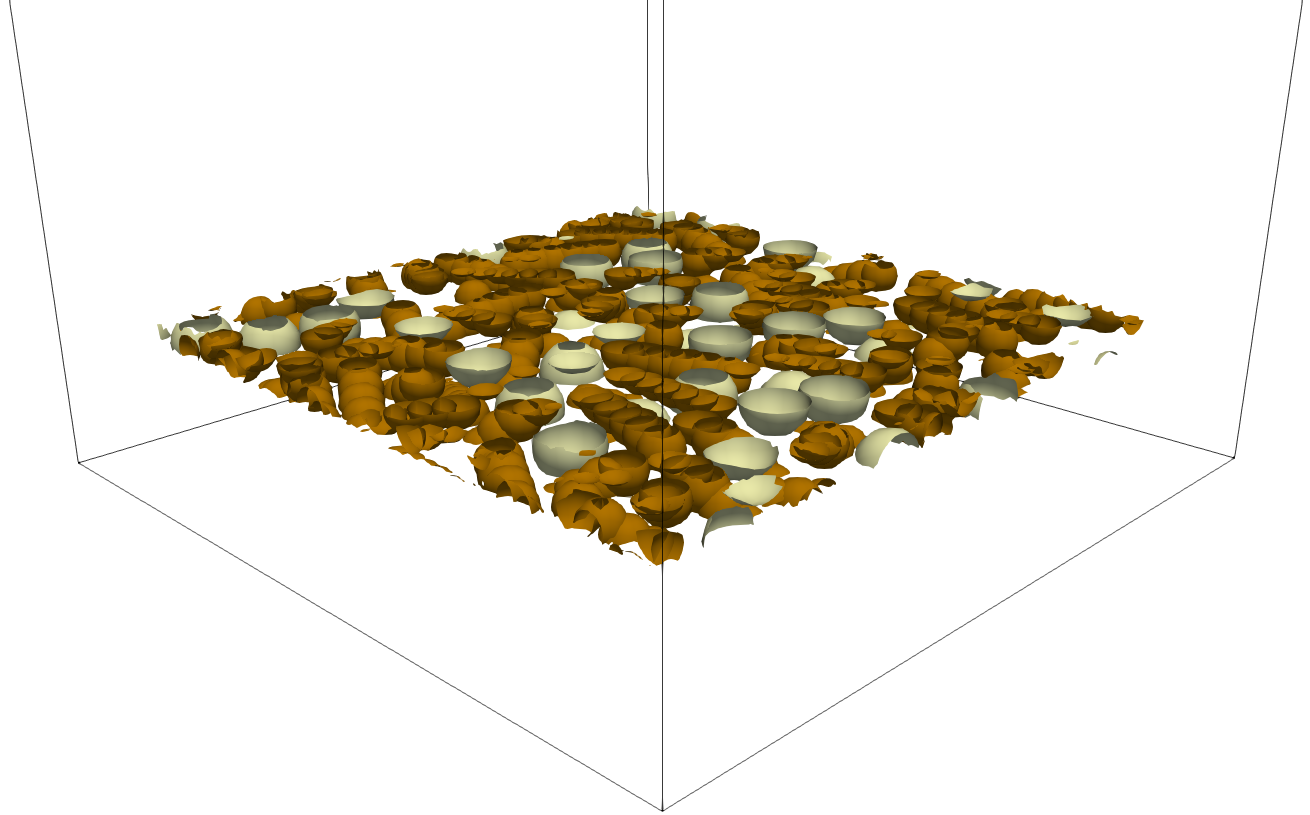}
            \caption{Sliced section}
            \label{fig:grav_dep_3d_d}
        \end{subfigure}
    \end{subfigure}
    \caption{3D visualizations of the deposition process with a 30\% pellet fraction. The settings differ between the visualizations. In the actual deposition, the woodchips and pellets are smaller and more numerous than shown here.}
    \label{fig:grav_dep_3d}
\end{figure}

To mimic the deposition process of the woodchips, we use Yade, a discrete-element method simulator \citep{Smilauer2023}.
The general approach involves initializing the woodchips as clumps of spheres, and dropping them inside a 3D box with gravity.
As we focus here on demonstrating our approach, we make the pragmatic choice of not calibrating against experiments.
Each woodchip is modeled as a rigid stack of connected spheres.
All woodchips are initialized with the same orientation $\theta$, as visualized in Figures \ref{fig:grav_dep_3d_a} \& \ref{fig:grav_dep_3d_b}, but a small random velocity is added to each to avoid an unrealistically aligned result.
As a result, most woodchips deviate to some extent from their initial orientation, but the remaining effect does lead to anisotropy.
The pellets are modeled as single spheres, with a radius 1.5 times as large as the spheres in the woodchips.
We control the \textit{pellet fraction}: the ratio between the volume of the pellets and the volume of the woodchips before deposition.
An example of a settled deposition is visualized in Figure \ref{fig:grav_dep_3d_c}.

\begin{figure}[htbp]
  \centering
  \begin{subfigure}[b]{0.242\textwidth}
      \centering
      \includegraphics[width=\textwidth]{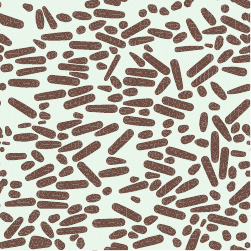}
      \caption{Pellet fraction: 0\%}
      \label{fig:grav_dep_mesh_fil_0.0}
  \end{subfigure}
  \hfill
  \begin{subfigure}[b]{0.242\textwidth}
      \centering
      \includegraphics[width=\textwidth]{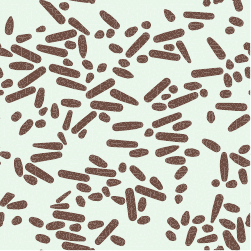}
      \caption{Pellet fraction: 30\%}
      \label{fig:grav_dep_mesh_fil_0.3}
  \end{subfigure}
    \hfill
  \begin{subfigure}[b]{0.242\textwidth}
      \centering
      \includegraphics[width=\textwidth]{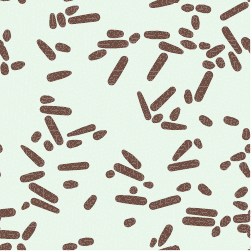}
      \caption{Pellet fraction: 60\%}
      \label{fig:grav_dep_mesh_fil_0.6}
  \end{subfigure}
    \hfill
  \begin{subfigure}[b]{0.242\textwidth}
      \centering
      \includegraphics[width=\textwidth]{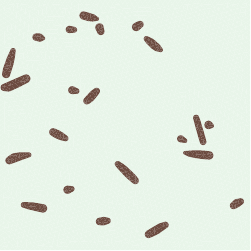}
      \caption{Pellet fraction: 90\%}
      \label{fig:grav_dep_mesh_fil_0.9}
  \end{subfigure}
  \caption{Generated 2D microstructures for various pellet fractions, all with an initial orientation of \ang{0}.}
  \label{fig:grav_dep_mesh_fil}
\end{figure}

Once the woodchips have settled, the configuration is sliced by a horizontal plane at a height of four times the radius of the spheres used to model the woodchips, as shown in Figure \ref{fig:grav_dep_3d_d}.
Only a central section of the box is included to avoid boundary effects of the 3D deposition.
Each intersected sphere is mapped to a disk in a 2D simulation (without gravity), whose radius equals the radius of the sphere's cross-section at the cutting height.
This 2D simulation has periodic boundary conditions, and a relaxation phase allows clumps that were overlapping other clumps at the boundary to untangle.
Then, a convex hull around each clump is drawn, before shrinking each hull to avoid overlap.
Finally, this geometry is meshed using Gmsh to create a periodic 2D mesh, as shown in Figure \ref{fig:grav_dep_mesh_fil}.
Only meshes from a single initial orientation are required since it can be accounted for by rotating the deformation and stress tensor as discussed in Section \ref{sec:param_and_data}; here we set the initial orientation to \ang{0}.

\begin{figure}[htbp]
    \centering
    \includegraphics[width=1.\textwidth]{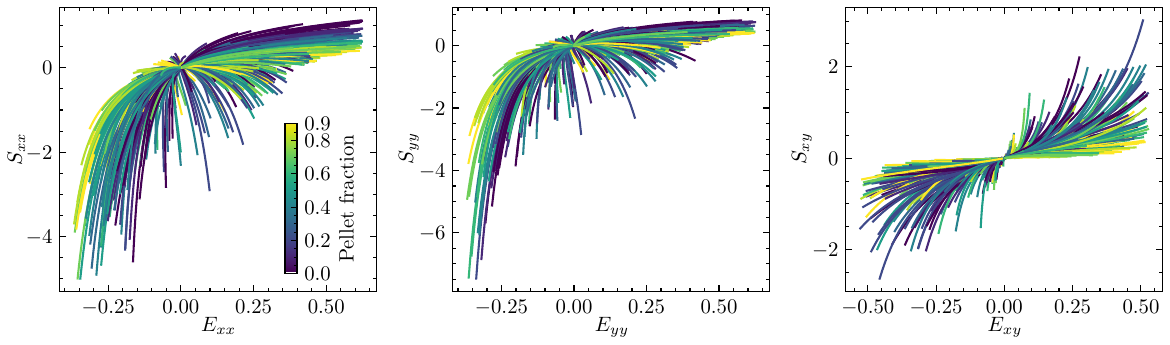}
    \caption{Visualization of the dataset created. The colormap shows the influence of different pellet fractions on the mechanical response.}
    \label{fig:grav_dep_dataset_fil_frac}
\end{figure}

This approach of taking a 2D slice from the 3D deposition does somewhat disconnect the manufacturing variable from the generated microstructure; the final model used is only indirectly related to the process that simulated it.
Ideally, the 3D deposition would be directly used, but we transfer it to 2D to reduce the computational cost of simulating the mechanical behavior of the microstructures.
The resulting meshes are subjected to monotonic loading in random directions to create a dataset.
Since the HyPRNN is conditioned on fewer parameters than before, we create a smaller dataset consisting of 512 samples, 256 used for training, and 128 each for validation and testing.
The dataset is visualized in Figure \ref{fig:grav_dep_dataset_fil_frac}.
An energy-based HyPRNN is trained on this dataset.

\subsection{Deformation optimization}
\begin{figure}[htbp]
    \centering
    \begin{subfigure}[t]{0.32\textwidth}
        \centering
        \includegraphics[width=\textwidth]{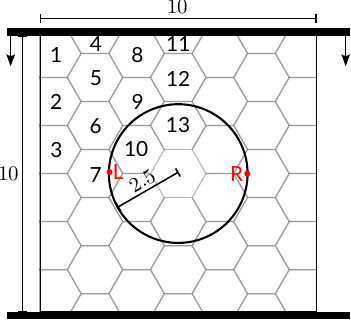}
        \caption{The thirteen parametrized regions are numbered.}
        \label{fig:grav_dep_optimization_overview}
    \end{subfigure}
    \hfill
    \begin{subfigure}[t]{0.325\textwidth}
        \centering
        \includegraphics[width=\textwidth]{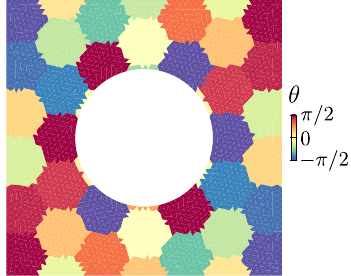}
        \caption{Optimized $\theta$}
        \label{fig:grav_dep_optimal_theta}
    \end{subfigure}
    \hfill
    \begin{subfigure}[t]{0.325\textwidth}
        \centering
        \includegraphics[width=\textwidth]{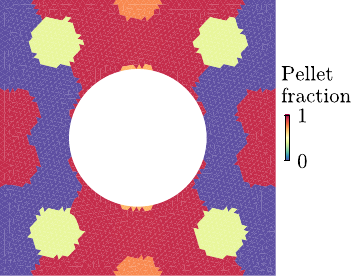}
        \caption{Optimized pellet fraction}
        \label{fig:grav_dep_optimal_frac}
    \end{subfigure}
    \caption{Simulation setup and results of the deformation optimization.}
    \label{fig:grav_dep_optimization}
\end{figure}

Next, we study whether the manufacturing variables can lead to a noticeable difference in a sample's deformation.
A compression test is simulated for a square sample with a hole; the setup is visualized in Figure \ref{fig:grav_dep_optimization_overview}.
The objective is to minimize the horizontal deformation of the hole, which we denote as bulging and quantify as: $u^R_x - u^L_x$, where $R$ and $L$ refer to the respective nodes in Figure \ref{fig:grav_dep_optimization_overview}.
The top plate is gradually lowered for a total deformation of 20\% of the sample height.
This loading is imposed with a contact model using the penalty method in order to replicate a compression test and allow for separation from the plate.
Since creating a continuous grading is challenging in practice, we instead partition the domain into hexagonal regions.
Using symmetry, this results in 13 regions for a total of 26 parameters to optimize.
The optimizer settings are identical to those of Section \ref{sec:grad_opt}.

\begin{figure}[htbp]
  \centering
  \begin{subfigure}[b]{0.48\textwidth}
      \centering
      \includegraphics[height=0.35\textheight]{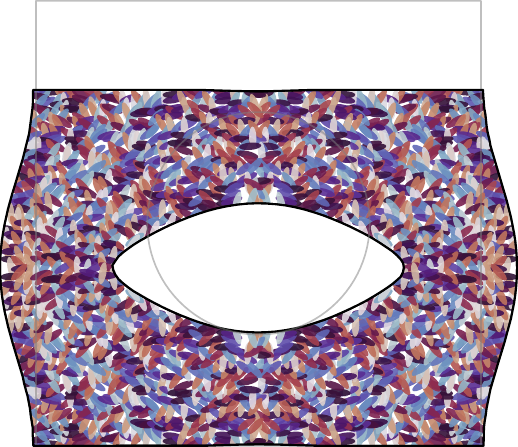}
      \caption{Baseline: bulge = 1.528.} 
      \label{fig:grav_baseline}
  \end{subfigure}
  \hfill
  \begin{subfigure}[b]{0.48\textwidth}
      \centering
      \includegraphics[height=0.35\textheight]{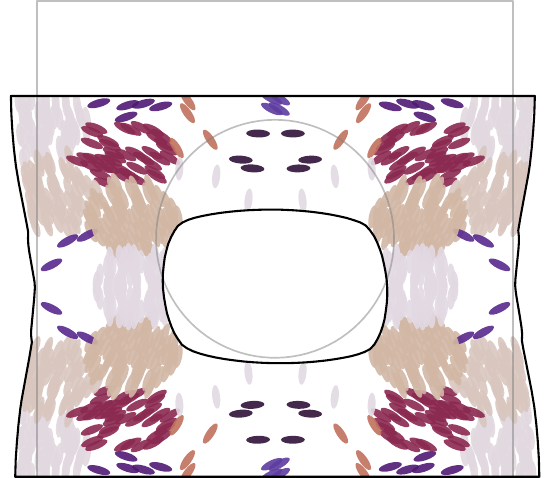}
      \caption{Optimized: bulge = -0.287.}    
      \label{fig:grav_dep_optimal}
  \end{subfigure}
  \caption{Deformed results from the baseline and optimized compression test. The ellipses represent both manufacturing variables: their sparsity depends (inversely) on the pellet fraction, and their orientation directly reflects their deposition orientation and deformation. Note that the size of the ellipses does not reflect the scale of the woodchips, for which we assume a separation of scales.}
  \label{fig:grav_dep_result}
\end{figure}

We compare the optimized result to a baseline case with 0\% pellet fraction and a random orientation in each element.
Figures \ref{fig:grav_dep_optimal_theta} and \ref{fig:grav_dep_optimal_frac} show the optimized woodchip orientation and pellet fraction.
The resulting microstructure distribution and deformations are visualized together in Figure \ref{fig:grav_dep_result}.
A clear difference is observed, with the baseline bulging significantly outward, whereas the optimized result bulges slightly inward.
The optimized pattern shows an X-like structure that forms bands from the corners of the sample to the L and R points.
Next to the bands there are parts with intermediate pellet fraction.
The optimized pattern of the orientation is far from trivial.
To verify the effectiveness of the orientation, we rerun the simulation with the optimized pellet fraction but a fixed orientation of $\theta$=\ang{90}, which leads to a worse result with a bulge of 0.689.
This example demonstrates how a HyPRNN can be used to find the manufacturing variables that lead to a desired mechanical behavior.

\section{Conclusion}

In order to develop a surrogate model for a range of different microstructures, we propose a HyPRNN: a PRNN conditioned on microstructural variables using a hypernetwork.
We compared two finite-strain HyPRNN formulations, namely a flexible nonlinear encoder and an energy-based formulation that maps strains linearly in logarithmic space.
Both formulations predict zero stress in the undeformed state and guarantee valid deformation tensors for the embedded material models.
The energy-based formulation is thermodynamically consistent.
The models performed similarly when trained on large datasets, but the stronger physics-based inductive bias of the energy HyPRNN resulted in lower errors on test samples and in an \fetwo{} simulation.
Material parameters such as the shear modulus can be varied directly in the embedded constitutive models without retraining, though less effectively with the nonlinear encoder.
Geometric properties, such as the volume fraction, are passed as inputs to the hypernetwork.
We handled orientations of the microstructure by training with one fixed orientation, and then pre- and post-processing the surrogate's inputs and outputs to account for the orientation.
This conditioned surrogate model enabled graded multiscale simulations.
For a simple 3-point bending experiment of a graded multiscale beam, the surrogate-based simulations reduced the computation time from 5900 seconds to 2 seconds, while maintaining high accuracy.
By performing a multiscale optimization, we showed the ability to find a material grading that significantly reduces peak stresses.

We circumvent the need for an additional model to parametrize microstructural geometries (e.g., a conditional variational autoencoder or a conditional denoising diffusion model) and instead condition directly on the manufacturing variables used to generate the microstructure.
Specifically, we conditioned the HyPRNN on inputs from a discrete-element simulation that mimics the deposition of woodchips and mycelium pellets.
The resulting model was used in an optimization loop to substantially alter the deformation pattern of a sample under compression.
The numerical examples in this work are limited to path-independent materials in 2D.
Given the success of PRNNs in capturing more complex constitutive behavior such as elasto-viscoplasticity and debonding, a natural extension is to test our HyPRNN formulation with those material models.
Furthermore, the material models and optimized structures should be calibrated and validated with physical experiments.
Overall, this work highlights the necessity of microstructure-aware data-driven surrogates for multiscale simulations.
Furthermore, we demonstrate the potential that precise control of microstructural variables can bring.

\section*{Acknowledgements}\label{sec:acknowledge}
JS, IR, and FM gratefully acknowledge the TU Delft AI Labs programme for enabling this work.
SS and KM are supported by the ERC Consolidator Grant, AM-IMATE, 101088968.
We also acknowledge Martin Lesueur and Winston Lindqwister for the insightful discussions.


%
\bibliographystyle{elsarticle-num}
\bibliography{reflib}

@article{Liu2019,
   author = {Zeliang Liu and C. T. Wu and M. Koishi},
   doi = {10.1016/j.cma.2018.09.020},
   issn = {00457825},
   journal = {Computer Methods in Applied Mechanics and Engineering},
   month = {3},
   pages = {1138-1168},
   publisher = {Elsevier B.V.},
   title = {A deep material network for multiscale topology learning and accelerated nonlinear modeling of heterogeneous materials},
   volume = {345},
   year = {2019}
}

@article{Gantenbein2022,
   author = {Silvan Gantenbein and Emanuele Colucci and Julian Käch and Etienne Trachsel and Fergal B. Coulter and Patrick A. Rühs and Kunal Masania and André R. Studart},
   doi = {10.1038/s41563-022-01429-5},
   issn = {1476-4660},
   issue = {1},
   journal = {Nature Materials 2022 22:1},
   month = {12},
   pages = {128-134},
   publisher = {Nature Publishing Group},
   title = {Three-dimensional printing of mycelium hydrogels into living complex materials},
   volume = {22},
   url = {https://www.nature.com/articles/s41563-022-01429-5},
   year = {2022}
}

@article{Islam2018,
   author = {M. R. Islam and G. Tudryn and R. Bucinell and L. Schadler and R. C. Picu},
   doi = {10.1007/s10853-018-2797-z},
   issn = {15734803},
   issue = {24},
   journal = {Journal of Materials Science},
   month = {12},
   pages = {16371-16382},
   publisher = {Springer New York LLC},
   title = {Mechanical behavior of mycelium-based particulate composites},
   volume = {53},
   year = {2018}
}

@article{MAMaia2024,
   author = {M.A. Maia and I.B.C.M. Rocha and D. Kovačević and F. P. van der Meer},
   doi = {10.1016/J.MECHMAT.2024.105145},
   issn = {0167-6636},
   journal = {Mechanics of Materials},
   month = {11},
   pages = {105145},
   publisher = {Elsevier},
   title = {Physically recurrent neural network for rate and path-dependent heterogeneous materials in a finite strain framework},
   volume = {198},
   year = {2024}
}

@article{Caruana1994,
   author = {Rich Caruana},
   journal = {Advances in Neural Information Processing Systems},
   title = {Learning Many Related Tasks at the Same Time with Backpropagation},
   volume = {7},
   year = {1994}
}

@article{Wei2025,
   author = {Ting-Ju Wei and Wen-Ning Wan and Chuin-Shan Chen},
   month = {4},
   title = {Deep Material Network: Overview, applications and current directions},
   url = {http://arxiv.org/abs/2504.12159},
   year = {2025}
}

@article{Li2024,
   author = {Tianyi Li},
   doi = {10.1016/J.CMA.2023.116687},
   issn = {0045-7825},
   journal = {Computer Methods in Applied Mechanics and Engineering},
   month = {2},
   pages = {116687},
   publisher = {North-Holland},
   title = {Micromechanics-informed parametric deep material network for physics behavior prediction of heterogeneous materials with a varying morphology},
   volume = {419},
   year = {2024}
}

@article{Storm2024,
   author = {J. Storm and I. B.C.M. Rocha and F. P. van der Meer},
   doi = {10.1016/J.CMA.2024.117001},
   issn = {0045-7825},
   journal = {Computer Methods in Applied Mechanics and Engineering},
   month = {7},
   pages = {117001},
   publisher = {North-Holland},
   title = {A microstructure-based graph neural network for accelerating multiscale simulations},
   volume = {427},
   year = {2024}
}

@article{Black2023,
   author = {Nolan Black and Ahmad R. Najafi},
   doi = {10.1007/S00158-022-03471-Y/FIGURES/26},
   issn = {16151488},
   issue = {1},
   journal = {Structural and Multidisciplinary Optimization},
   month = {1},
   pages = {1-25},
   publisher = {Springer Science and Business Media Deutschland GmbH},
   title = {Deep neural networks for parameterized homogenization in concurrent multiscale structural optimization},
   volume = {66},
   url = {https://link.springer.com/article/10.1007/s00158-022-03471-y},
   year = {2023}
}

@article{Rao2020,
   author = {Chengping Rao and Yang Liu},
   doi = {10.1016/J.COMMATSCI.2020.109850},
   issn = {0927-0256},
   journal = {Computational Materials Science},
   month = {11},
   pages = {109850},
   publisher = {Elsevier},
   title = {Three-dimensional convolutional neural network (3D-CNN) for heterogeneous material homogenization},
   volume = {184},
   year = {2020}
}

@article{Fuhg2024,
   author = {Jan N. Fuhg and Govinda Anantha Padmanabha and Nikolaos Bouklas and Bahador Bahmani and Wai Ching Sun and Nikolaos N. Vlassis and Moritz Flaschel and Pietro Carrara and Laura De Lorenzis},
   doi = {10.1007/S11831-024-10196-2},
   isbn = {0123456789},
   issn = {1886-1784},
   issue = {3},
   journal = {Archives of Computational Methods in Engineering 2024 32:3},
   month = {11},
   pages = {1841-1883},
   publisher = {Springer},
   title = {A Review on Data-Driven Constitutive Laws for Solids},
   volume = {32},
   url = {https://link.springer.com/article/10.1007/s11831-024-10196-2},
   year = {2024}
}

@article{Kovcs2025,
   author = {N. Kovács and I. B. C. M. Rocha and F. P. van der Meer and C. Furtado and P. P. Camanho},
   title = {Uncertainty Quantification in Multiscale Modeling of Polymer Composite Materials Using Physically Recurrent Neural Networks},
   url = {http://arxiv.org/abs/2504.11625},
   year = {2025}
}

@article{Pitz2024,
   author = {Emil Pitz and Kishore Pochiraju},
   doi = {10.1016/J.ENGAPPAI.2024.108622},
   issn = {0952-1976},
   journal = {Engineering Applications of Artificial Intelligence},
   month = {8},
   pages = {108622},
   publisher = {Pergamon},
   title = {A neural network transformer model for composite microstructure homogenization},
   volume = {134},
   year = {2024}
}

@article{Heidenreich2024,
   author = {Julian N. Heidenreich and Colin Bonatti and Dirk Mohr},
   doi = {10.1002/NME.7357},
   issn = {1097-0207},
   issue = {1},
   journal = {International Journal for Numerical Methods in Engineering},
   month = {1},
   pages = {e7357},
   publisher = {John Wiley \& Sons, Ltd},
   title = {Transfer learning of recurrent neural network-based plasticity models},
   volume = {125},
   url = {https://onlinelibrary.wiley.com/doi/full/10.1002/nme.7357 https://onlinelibrary.wiley.com/doi/abs/10.1002/nme.7357 https://onlinelibrary.wiley.com/doi/10.1002/nme.7357},
   year = {2024}
}

@article{Ghane2024,
   author = {Ehsan Ghane and Martin Fagerström and Mohsen Mirkhalaf},
   doi = {10.1016/J.EUROMECHSOL.2024.105378},
   issn = {0997-7538},
   journal = {European Journal of Mechanics - A/Solids},
   month = {9},
   pages = {105378},
   publisher = {Elsevier Masson},
   title = {Recurrent neural networks and transfer learning for predicting elasto-plasticity in woven composites},
   volume = {107},
   year = {2024}
}

@misc{Finn2017,
   author = {Chelsea Finn and Pieter Abbeel and Sergey Levine},
   issn = {2640-3498},
   month = {7},
   pages = {1126-1135},
   publisher = {PMLR},
   title = {Model-Agnostic Meta-Learning for Fast Adaptation of Deep Networks},
   url = {https://proceedings.mlr.press/v70/finn17a.html},
   year = {2017}
}

@book{Miyamoto2013,
   author = {Yoshinari and Kaysser, WA and Rabin, BH and Kawasaki, Akira and Ford, Rene\{\'e\} G Miyamoto},
   publisher = {Springer Science \& Business Media},
   title = {Functionally Graded Materials: Design, Processing and Applications},
   volume = {5},
   year = {2013}
}

@article{Zheng2024,
   author = {Li Zheng and Dennis M. Kochmann and Siddhant Kumar},
   doi = {10.1016/j.eml.2024.102243},
   issn = {23524316},
   journal = {Extreme Mechanics Letters},
   month = {11},
   publisher = {Elsevier Ltd},
   title = {HyperCAN: Hypernetwork-driven deep parameterized constitutive models for metamaterials},
   volume = {72},
   year = {2024}
}

@article{Jailin2025,
   author = {C. Jailin and E. Baranger},
   doi = {10.1016/J.CMA.2025.118188},
   issn = {0045-7825},
   journal = {Computer Methods in Applied Mechanics and Engineering},
   month = {10},
   pages = {118188},
   publisher = {North-Holland},
   title = {Material-embedding Physics-Augmented Neural Networks: A first application to constitutive law parameterization},
   volume = {445},
   year = {2025}
}

@article{Mozaffar2019,
   author = {M. Mozaffar and R. Bostanabad and W. Chen and K. Ehmann and Jian Cao and M. A. Bessa},
   doi = {10.1073/PNAS.1911815116/-/DCSUPPLEMENTAL.Y},
   issn = {10916490},
   issue = {52},
   journal = {Proceedings of the National Academy of Sciences of the United States of America},
   month = {12},
   pages = {26414-26420},
   pmid = {31843918},
   publisher = {National Academy of Sciences},
   title = {Deep learning predicts path-dependent plasticity},
   volume = {116},
   url = {https://github.com/mabessa/F3DAM.yThisarticlecontainssupportinginformationonlineathttps://www.pnas.org/lookup/suppl/www.pnas.org/cgi/doi/10.1073/pnas.1911815116},
   year = {2019}
}

@article{Maia2023,
   author = {M. A. Maia and I. B.C.M. Rocha and P. Kerfriden and F. P. van der Meer},
   doi = {10.1016/J.CMA.2023.115934},
   issn = {0045-7825},
   journal = {Computer Methods in Applied Mechanics and Engineering},
   month = {3},
   pages = {115934},
   publisher = {North-Holland},
   title = {Physically recurrent neural networks for path-dependent heterogeneous materials: Embedding constitutive models in a data-driven surrogate},
   volume = {407},
   year = {2023}
}

@article{Wang2024,
   author = {Hao Wang and Jie Tao and Zhangyu Wu and Kathrin Weiland and Zuankai Wang and Kunal Masania and Bin Wang},
   doi = {10.1002/ADVS.202309370},
   issn = {21983844},
   issue = {24},
   journal = {Advanced Science},
   month = {6},
   pmid = {38477443},
   publisher = {John Wiley and Sons Inc},
   title = {Fabrication of Living Entangled Network Composites Enabled by Mycelium},
   volume = {11},
   year = {2024}
}

@article{Islam2018b,
   author = {M. R. Islam and G. Tudryn and R. Bucinell and L. Schadler and R. C. Picu},
   doi = {10.1016/J.MATDES.2018.09.046},
   issn = {0264-1275},
   journal = {Materials \& Design},
   month = {12},
   pages = {549-556},
   publisher = {Elsevier},
   title = {Stochastic continuum model for mycelium-based bio-foam},
   volume = {160},
   year = {2018}
}

@article{Chandrasekhar2023,
   author = {Aaditya Chandrasekhar and Saketh Sridhara and Krishnan Suresh},
   doi = {10.1016/J.ADVENGSOFT.2022.103359},
   issn = {0965-9978},
   journal = {Advances in Engineering Software},
   month = {1},
   pages = {103359},
   publisher = {Elsevier},
   title = {Graded multiscale topology optimization using neural networks},
   volume = {175},
   year = {2023}
}

@article{Kim2021,
   author = {Cheolwoong Kim and Jaewook Lee and Jeonghoon Yoo},
   doi = {10.1016/J.CMA.2021.114158},
   issn = {0045-7825},
   journal = {Computer Methods in Applied Mechanics and Engineering},
   month = {12},
   pages = {114158},
   publisher = {North-Holland},
   title = {Machine learning-combined topology optimization for functionary graded composite structure design},
   volume = {387},
   year = {2021}
}

@article{Zheng2021,
   author = {Li Zheng and Siddhant Kumar and Dennis M. Kochmann},
   doi = {10.1016/J.CMA.2021.113894},
   issn = {0045-7825},
   journal = {Computer Methods in Applied Mechanics and Engineering},
   month = {9},
   pages = {113894},
   publisher = {North-Holland},
   title = {Data-driven topology optimization of spinodoid metamaterials with seamlessly tunable anisotropy},
   volume = {383},
   year = {2021}
}

@article{Islam2017,
   author = {M. R. Islam and G. Tudryn and R. Bucinell and L. Schadler and R. C. Picu},
   doi = {10.1038/s41598-017-13295-2},
   issn = {2045-2322},
   issue = {1},
   journal = {Scientific Reports 2017 7:1},
   month = {10},
   pages = {13070-},
   pmid = {29026133},
   publisher = {Nature Publishing Group},
   title = {Morphology and mechanics of fungal mycelium},
   volume = {7},
   url = {https://www.nature.com/articles/s41598-017-13295-2},
   year = {2017}
}

@article{baratta2023dolfinx,
   author = {Igor A. Baratta and Joseph P. Dean and Jørgen S. Dokken and Michal; University of Luxembourg; Rafinex Sarl HABERA and Jack; University of Luxembourg > Faculty of Science, Technology and Medicine (FSTM) > Department of Engineering (DoE) HALE and Chris N. Richardson and Marie E. Rognes and Matthew W. Scroggs and Nathan Sime and Garth N. Wells},
   doi = {10.5281/ZENODO.10447666},
   journal = {Zenodo},
   month = {12},
   title = {DOLFINx: The next generation FEniCS problem solving environment},
   url = {https://orbilu.uni.lu/handle/10993/59327},
   year = {2025}
}

@article{Bayat2008,
   author = {Mehdi Bayat and M. Saleem and B. B. Sahari and A. M.S. Hamouda and E. Mahdi},
   doi = {10.1016/J.MECHRESCOM.2008.02.007},
   issn = {0093-6413},
   issue = {5},
   journal = {Mechanics Research Communications},
   month = {7},
   pages = {283-309},
   publisher = {Pergamon},
   title = {Analysis of functionally graded rotating disks with variable thickness},
   volume = {35},
   year = {2008}
}

@article{Durodola2000,
   author = {J. F. Durodola and O. Attia},
   doi = {10.1016/S0266-3538(99)00197-9},
   issn = {0266-3538},
   issue = {7},
   journal = {Composites Science and Technology},
   month = {5},
   pages = {987-995},
   publisher = {Elsevier},
   title = {Deformation and stresses in functionally graded rotating disks},
   volume = {60},
   year = {2000}
}

@article{Abdalla2020,
   author = {Hassan Mohamed Abdelalim Abdalla and Daniele Casagrande and Luciano Moro},
   doi = {10.1177/0309324720904793/ASSET/9D49D730-B9F7-4A4A-8CD7-9F591A582618/ASSETS/IMAGES/LARGE/10.1177_0309324720904793-FIG14.JPG},
   issn = {20413130},
   issue = {5-6},
   journal = {Journal of Strain Analysis for Engineering Design},
   month = {8},
   pages = {159-171},
   publisher = {SAGE Publications Ltd},
   title = {Thermo-mechanical analysis and optimization of functionally graded rotating disks},
   volume = {55},
   url = {https://journals.sagepub.com/doi/full/10.1177/0309324720904793},
   year = {2020}
}

@article{Hansen2001,
   author = {N. Hansen and A. Ostermeier},
   doi = {10.1162/106365601750190398},
   issn = {10636560},
   issue = {2},
   journal = {Evolutionary computation},
   pages = {159-195},
   pmid = {11382355},
   title = {Completely derandomized self-adaptation in evolution strategies.},
   volume = {9},
   year = {2001}
}

@article{Smilauer2023,
   author = {Vaclav Smilauer and Vasileios Angelidakis and Emanuele Catalano and Robert Caulk and Bruno Chareyre and William Chevremont and Sergei Dorofeenko and Jerome Duriez and Nolan Dyck and Jan Elias and others},
   journal = {arXiv preprint arXiv:2301.00611},
   title = {Yade documentation},
   year = {2023}
}

@misc{Bradbury2018,
   author = {James Bradbury and Roy Frostig and Peter Hawkins and Matthew James Johnson and Yash Katariya and Chris Leary and Dougal Maclaurin and George Necula and Adam Paszke and Jake VanderPlas and Skye Wanderman-Milne and Qiao Zhang},
   title = {JAX: composable transformations of Python+NumPy programs},
   url = {http://github.com/jax-ml/jax},
   year = {2018}
}

@misc{Bleyer2024,
   author = {Jeremy Bleyer},
   doi = {10.5281/zenodo.13882183},
   month = {10},
   publisher = {Zenodo},
   title = {dolfinx\_materials: A Python package for advanced
material modelling},
   url = {https://doi.org/10.5281/zenodo.13882183},
   year = {2024}
}

@article{Kingma2014,
   author = {Diederik P. Kingma and Jimmy Lei Ba},
   journal = {3rd International Conference on Learning Representations, ICLR 2015 - Conference Track Proceedings},
   month = {12},
   publisher = {International Conference on Learning Representations, ICLR},
   title = {Adam: A Method for Stochastic Optimization},
   url = {https://arxiv.org/abs/1412.6980v9},
   year = {2014}
}

@article{Geuzaine2009,
   author = {Christophe Geuzaine and Jean François Remacle},
   doi = {10.1002/NME.2579},
   issn = {1097-0207},
   issue = {11},
   journal = {International Journal for Numerical Methods in Engineering},
   month = {9},
   pages = {1309-1331},
   publisher = {John Wiley \& Sons, Ltd},
   title = {Gmsh: A 3-D finite element mesh generator with built-in pre- and post-processing facilities},
   volume = {79},
   url = {https://onlinelibrary.wiley.com/doi/full/10.1002/nme.2579 https://onlinelibrary.wiley.com/doi/abs/10.1002/nme.2579 https://onlinelibrary.wiley.com/doi/10.1002/nme.2579},
   year = {2009}
}

@article{Zhang2023,
   author = {Mingchang Zhang and Zhenxin Zhang and Runhua Zhang and Yao Peng and Mingzhi Wang and Jinzhen Cao},
   doi = {10.1016/J.COMPOSITESB.2023.111003},
   issn = {1359-8368},
   journal = {Composites Part B: Engineering},
   month = {11},
   pages = {111003},
   publisher = {Elsevier},
   title = {Lightweight, thermal insulation, hydrophobic mycelium composites with hierarchical porous structure: Design, manufacture and applications},
   volume = {266},
   year = {2023}
}

@article{Schyck2026,
   author = {Sarah Schyck and Mark Ablonczy and Sourav Patranabish and Kunal Masania},
   doi = {10.1002/ADFM.202530836},
   issn = {1616-3028},
   issue = {38},
   journal = {Advanced Functional Materials},
   month = {5},
   pages = {e30836},
   publisher = {John Wiley \& Sons, Ltd},
   title = {Shaping of Biohybrid Functional Living Materials},
   volume = {36},
   url = {https://onlinelibrary.wiley.com/doi/full/10.1002/adfm.202530836 https://onlinelibrary.wiley.com/doi/abs/10.1002/adfm.202530836 https://advanced.onlinelibrary.wiley.com/doi/10.1002/adfm.202530836},
   year = {2026}
}

@article{McBee2021,
   author = {Ross M. McBee and Matt Lucht and Nikita Mukhitov and Miles Richardson and Tarun Srinivasan and Dechuan Meng and Haorong Chen and Andrew Kaufman and Max Reitman and Christian Munck and Damen Schaak and Christopher Voigt and Harris H. Wang},
   doi = {10.1038/s41563-021-01123-y},
   issn = {1476-4660},
   issue = {4},
   journal = {Nature Materials 2021 21:4},
   month = {12},
   pages = {471-478},
   pmid = {34857911},
   publisher = {Nature Publishing Group},
   title = {Engineering living and regenerative fungal–bacterial biocomposite structures},
   volume = {21},
   url = {https://www.nature.com/articles/s41563-021-01123-y},
   year = {2021}
}

@article{Nettersheim2024,
   author = {I. H.M.S. Nettersheim and N. S.Guevara Sotelo and J. C. Verdonk and K. Masania},
   doi = {10.1016/J.COMPSCITECH.2024.110758},
   issn = {0266-3538},
   journal = {Composites Science and Technology},
   month = {9},
   pages = {110758},
   publisher = {Elsevier},
   title = {Engineered living composite materials},
   volume = {256},
   year = {2024}
}

@article{Chafia2025,
   author = {Zakaria Chafia and Julien Yvonnet and Jérémy Bleyer},
   doi = {10.1016/J.CMA.2025.117747},
   issn = {0045-7825},
   journal = {Computer Methods in Applied Mechanics and Engineering},
   month = {3},
   pages = {117747},
   publisher = {North-Holland},
   title = {A Data-Driven-based homogenization method to simulate the anisotropic damage of brittle heterogeneous structures},
   volume = {437},
   year = {2025}
}

@article{Ammendolea2026,
   author = {Domenico Ammendolea and Fabrizio Greco and Lorenzo Leonetti and Arturo Pascuzzo},
   doi = {10.1016/j.compstruct.2025.119869},
   issn = {02638223},
   journal = {Composite Structures},
   month = {2},
   publisher = {Elsevier Ltd},
   title = {Prediction of anisotropic damage evolution in nacre-inspired composites by using a data-driven nonlinear homogenization approach},
   volume = {377},
   year = {2026}
}

@article{Linka2021,
   author = {Kevin Linka and Markus Hillgärtner and Kian P. Abdolazizi and Roland C. Aydin and Mikhail Itskov and Christian J. Cyron},
   doi = {10.1016/j.jcp.2020.110010},
   issn = {10902716},
   journal = {Journal of Computational Physics},
   month = {3},
   publisher = {Academic Press Inc.},
   title = {Constitutive artificial neural networks: A fast and general approach to predictive data-driven constitutive modeling by deep learning},
   volume = {429},
   year = {2021}
}

@article{Klein2022,
   author = {Dominik K. Klein and Mauricio Fernández and Robert J. Martin and Patrizio Neff and Oliver Weeger},
   doi = {10.1016/j.jmps.2021.104703},
   issn = {00225096},
   journal = {Journal of the Mechanics and Physics of Solids},
   month = {2},
   publisher = {Elsevier Ltd},
   title = {Polyconvex anisotropic hyperelasticity with neural networks},
   volume = {159},
   year = {2022}
}

@article{Maia2026,
   author = {M. A. Maia and K. A. Meyer and A. M. C. M. van Gils and I. B. C. M. Rocha and F. P. van der Meer},
   month = {10},
   title = {On architectural choices for interpretability and thermodynamic consistency in Physically Recurrent Neural Networks in the low-data regime},
   url = {https://arxiv.org/abs/2610.04067v1},
   year = {2026}
}

\appendix

\section{Hyperparameter configurations}\label{app:hyperparameters}
In Table \ref{tab:sweep_options} we present the hyperparameters used to create the learning curves and compare the model performance.
A wider range was explored in earlier prototyping.
We did not perform a full grid sweep over the possible settings for each model, but followed a more manual iterative process.
For example, for the neural network, we found that ReLU performed worse than the other activations, and therefore did not try all model sizes with it.
For the hypernetwork, we further considered an alternative setup that scaled the hidden layer sizes depending on the number of its outputs (the number of encoder weights), but found no gain.

\begin{table}[htbp]
\centering
\caption{Hyperparameter sweep options for the various models. The neural network hidden layer sizes are specified by the number of neurons of the hidden layer, excluding the input and output layers. Scalar material point values reflect the number of mycelium material points, the [6,3] array reflects 6 mycelium material points, and 3 woodchip material points.}
\label{tab:sweep_options}
\begin{tabular}{llll}
\toprule
Hyperparameter & HyPRNN Nonlinear & HyPRNN Energy & NN \\
\midrule
Material points           & 2, 3, 6, 12, 24, [6,3]  & 6, 12, 18, 24, [12,6]     & -- \\
Hypernetwork activations   & sigmoid, ReLU           & sigmoid, ReLU     & --            \\
\multirow{2}{*}{Hypernetwork hidden sizes}
                & \multirow{2}{*}{(8), (8,8,8)} & (4), (8), (12), & \multirow{2}{*}{--} \\
                & & (24), (8,8) & \\
Encoder activations       & softplus, sigmoid, ReLU & --     & --                 \\
\multirow{2}{*}{Encoder hidden sizes}
                & (8), (8,8,8), (16),  & \multirow{2}{*}{--} & \multirow{2}{*}{--}                 \\
                & (16,16,16) & &    \\
NN activations            & -- & -- & sigmoid, tanh, ReLU                  \\
\multirow{3}{*}{NN hidden sizes}
                          &  &  & (8,8), (16,16), (16,16,16), \\
                          & -- & -- & (32,32,32), (64,64), (64,64,64), \\
                          &  &  & (64,64,64,64), (128,128,128) \\\bottomrule
\end{tabular}
\end{table}

\begin{figure}[htbp]
    \centering
    \includegraphics[width=0.4\textwidth]{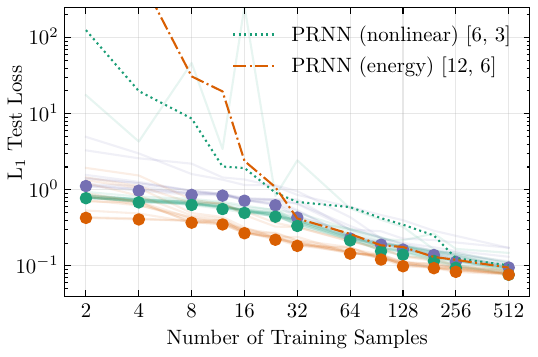}
    \caption{Learning curves on $\mathcal{D}^{all}$ with two models that embed woodchip material points highlighted.}
    \label{fig:learn_curve_vfrac_ratio_6m3}
\end{figure}

In Figure \ref{fig:learn_curve_vfrac_ratio_6m3} we highlight learning curves for models with woodchip material points, one for the nonlinear and one for the energy PRNN.
The woodchip material points cause significantly worse losses in the low-data regime.
Given sufficient data, the models learn to make good predictions, but are still among the worst relative to other configurations.

\begin{figure}[htbp]
    \centering
    \begin{subfigure}[b]{0.325\textwidth}
        \centering
        \includegraphics[width=\textwidth]{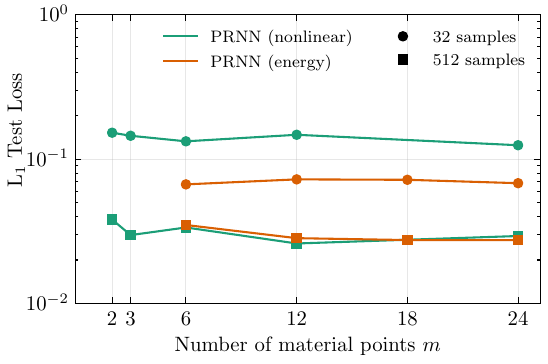}
        \caption{$\mathcal{D}^{mat}$}
        \label{fig:learn_curve_matpoints_mu}
    \end{subfigure}
    \hfill
    \begin{subfigure}[b]{0.325\textwidth}
        \centering
        \includegraphics[width=\textwidth]{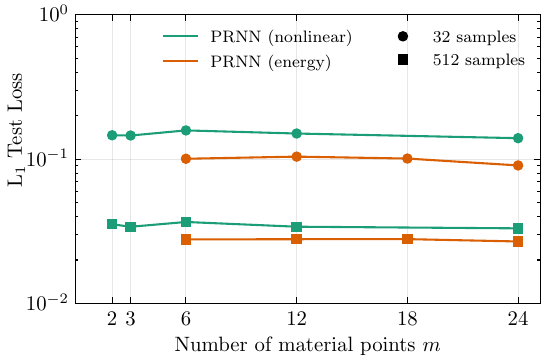}
        \caption{$\mathcal{D}^{geo}$}
        \label{fig:learn_curve_matpoints_vfrac_ratio}
    \end{subfigure}
    \hfill
    \begin{subfigure}[b]{0.325\textwidth}
        \centering
        \includegraphics[width=\textwidth]{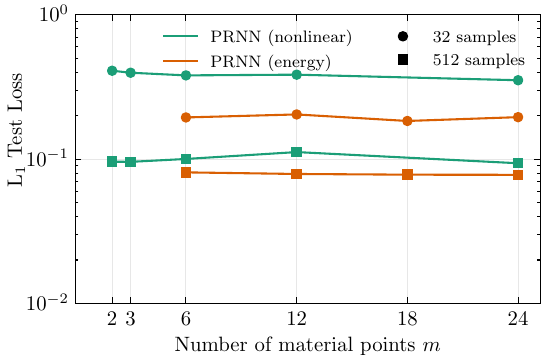}
        \caption{$\mathcal{D}^{all}$}
        \label{fig:learn_curve_matpoints_all}
    \end{subfigure}
    \caption{Influence of the number of fictitious material points on the test loss, given either 32 or 512 training samples. Results are averaged over 10 training runs per configuration.}
    \label{fig:learn_curve_matpoints}
\end{figure}

The influence of the number of material points on the test loss is visualized in Figure \ref{fig:learn_curve_matpoints}.
The performance is remarkably similar across a wide range of values.
This insensitivity is beneficial for model selection.

\section{\fetwo{} validation stress paths}\label{app:fe2val}

\begin{figure}[htbp]
    \centering
    \includegraphics[width=.5\textwidth]{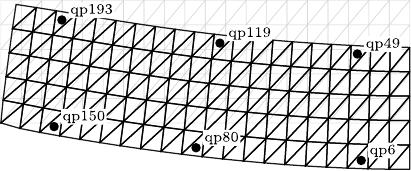}
    \caption{Selected quadrature points.}
    \label{fig:app_qp_points}
\end{figure}

\begin{figure}[htbp]
    \centering
    \includegraphics[width=1.\textwidth]{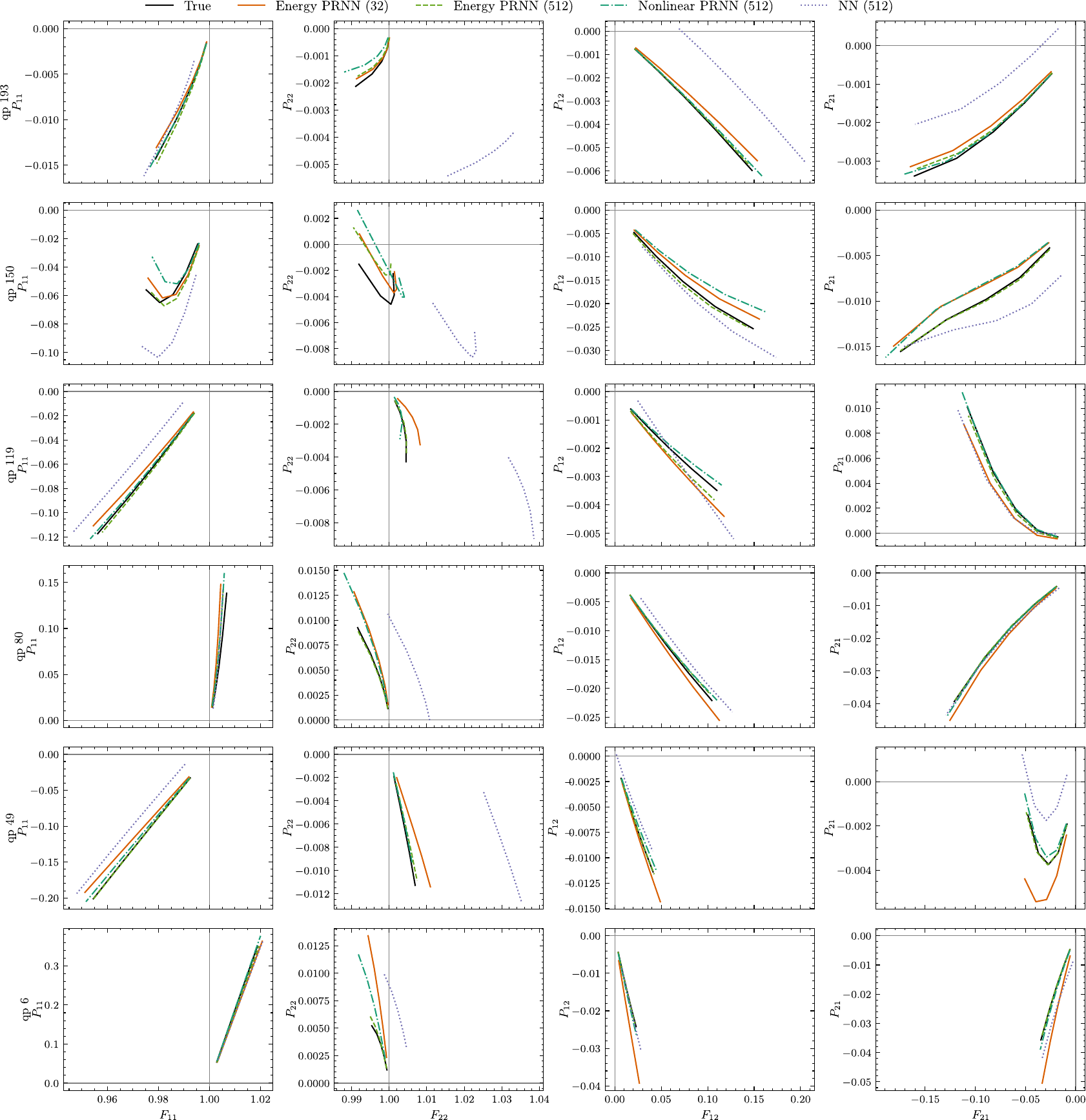}
    \caption{Stress strain paths for selected quadrature points in the 3-point bending beam.}
    \label{fig:app_val_stress_strain_paths}
\end{figure}

In Figure \ref{fig:app_val_stress_strain_paths} we plot the stress strain paths of several quadrature points from during the simulations in Section \ref{sec:fe2_validation}.
The selected quadrature points are shown in Figure \ref{fig:app_qp_points}.
We omit the results of the NN (32) surrogate, as its errors are high, and including its paths requires a substantial increase of the axes bounds, making the other paths more challenging to compare.
This figure clearly demonstrates how the NN is not constrained to have zero stresses in the undeformed state, showing a significant bias.
All PRNN models follow the ground truth response.

\end{document}